\documentclass[10pt,twocolumn,letterpaper]{article}

\usepackage{wacv}
\usepackage{times}
\usepackage{epsfig}
\usepackage{graphicx}
\usepackage{amsmath}
\usepackage{amssymb}

\def\wacvPaperID{370} 

\wacvalgorithmstrack   

\wacvfinalcopy 

\ifwacvfinal
\def\assignedStartPage{1} 
\fi

\ifwacvfinal
\usepackage[breaklinks=true,bookmarks=false,hypertexnames=false]{hyperref}
\else
\usepackage[pagebackref=true,breaklinks=true,colorlinks,bookmarks=false]{hyperref}
\fi

\ifwacvfinal
\else
\fi

\newcommand{\myparagraph}[1]{\vspace{0pt}{\bf #1}}

\usepackage[T1]{fontenc}
\usepackage[table,RGB]{xcolor}
\newcommand{\anna}[1]{\textcolor{blue}{Anna: #1}}

\usepackage{amsmath} 
\usepackage{pifont}
\definecolor{darkgreen}{rgb}{0, 0.4, 0}
\definecolor{darkred}{rgb}{0.7, 0, 0}
\definecolor{darkblue}{rgb}{0.0, 0.0, 0.7}
\newcommand{\cmark}{\textcolor{darkgreen}{\ding{51}}}%
\newcommand{\xmark}{\textcolor{darkred}{\ding{55}}}%
\usepackage{multirow}
\usepackage{textcomp}
\usepackage{graphicx}
\usepackage{bbm}
\usepackage{etoolbox}
\usepackage{capt-of}

\usepackage{xr}

\begin{document}

\title{One-Shot Synthesis of Images and Segmentation Masks}


\author{Vadim Sushko$^\text{1}$~~~~Dan Zhang$^\text{1,2}$~~~~Juergen Gall$^\text{3}$~~~~Anna Khoreva$^\text{1,2}$\\
	$^\text{1}$Bosch Center for Artificial Intelligence~~~~~~$^\text{2}$University of T{\"u}bingen~~~~~~$^\text{3}$University of Bonn\\
	{\tt\small \{vadim.sushko,dan.zhang2,anna.khoreva\}@bosch.com,~gall@iai.uni-bonn.de}
}

\makeatletter
\apptocmd\@maketitle{{\myfigure{}\par}}{}{}
\makeatother

\newcommand\myfigure{%
	\vspace{-1.5em}
	\begin{centering}
		\setlength{\tabcolsep}{0.1em}
		\renewcommand{\arraystretch}{1.0}
		\par\end{centering}
		\begin{centering}
		~~\includegraphics[width=0.99\linewidth]{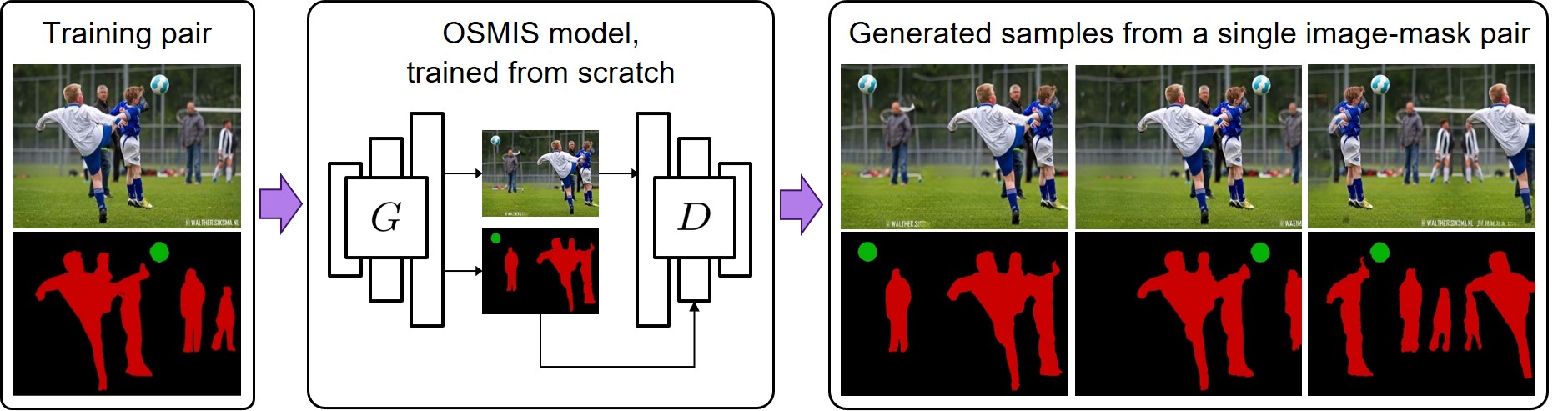} 
		\par\end{centering}
		\vspace{-0.0em}
		\captionof{figure}{We introduce a new task of generating new images and their segmentation masks from a single training pair, without access to any pre-training data. Under this challenging regime, our proposed GAN model (OSMIS) achieves a synthesis of a high structural diversity, preserving the photorealism of original images and a precise alignment of produced segmentation masks to the generated content. 
		}
		\label{fig:teaser}
		\vspace{1.5em}
}

\maketitle


\begin{abstract}
	\vspace{-0.5cm}
	Joint synthesis of images and segmentation masks with generative adversarial networks (GANs) is promising to reduce the effort needed for collecting image data with pixel-wise annotations. However, to learn high-fidelity image-mask synthesis, existing GAN approaches first need a pre-training phase requiring large amounts of image data,
	which limits their utilization 
	in restricted image domains.
	In this work, we take a step to reduce this limitation, introducing the task of one-shot image-mask synthesis. We aim to generate diverse images and their segmentation masks given only a single labelled example, and assuming, contrary to previous models, no access to any pre-training data. 
	To this end, inspired by the recent architectural developments of single-image GANs, we introduce our OSMIS model which enables the synthesis of segmentation masks that are precisely aligned to the generated images in the one-shot regime. Besides achieving the high fidelity of generated masks, OSMIS outperforms state-of-the-art single-image GAN models in image synthesis quality and diversity.
	In addition, despite not using any additional data, OSMIS demonstrates an impressive ability to serve as a source of useful data augmentation for one-shot segmentation applications, providing performance gains that are complementary to standard data augmentation techniques.
	Code is available at \url{https://github.com/boschresearch/one-shot-synthesis}.
\end{abstract}
\vspace{-3.0ex}

\section{Introduction}
\label{sec:introduction}

Deep neural networks have been shown powerful at solving various segmentation problems in computer vision \cite{chen2018encoder, he2017mask, kirillov2019panoptic, pont20172017, nilsson2018semantic, wang2019panet}.
The success of these segmentation models strongly relies on the availability of a large-scale collection of labelled data for training. Nevertheless, annotation of a large dataset is not always feasible in practice due to a very high cost of manual labelling of segmentation masks \cite{caesar2018coco}. For example, accurately labelling a single image with many objects can take more than 30 minutes \cite{zhang2021datasetgan}. Therefore, diminishing the human effort required for obtaining diverse and precisely aligned image-mask data is an important problem for many practical applications.


\begin{figure*}[t]
	\begin{centering}
		\setlength{\tabcolsep}{0.0em}
		\par\end{centering}
	\renewcommand{\arraystretch}{1.00}
	\centering
	\vspace{-0.5ex}
	\begin{tabular}{@{\hskip 0.02in}c@{\hskip 0.06in}c@{\hskip 0.03in}c@{\hskip 0.06in}c@{\hskip 0.03in}c@{\hskip 0.06in}c@{\hskip 0.03in}c@{\hskip 0.03in}c@{\hskip 0.03in}c}
	\small Training pair & \multicolumn{2}{c}{\small SemanticGAN \cite{li2021semantic}} &  \multicolumn{2}{c}{DatasetGAN \cite{zhang2021datasetgan}} & \multicolumn{3}{c}{\small OSMIS}
	\tabularnewline 
	\includegraphics[width=0.118\linewidth, height=0.13\textheight]{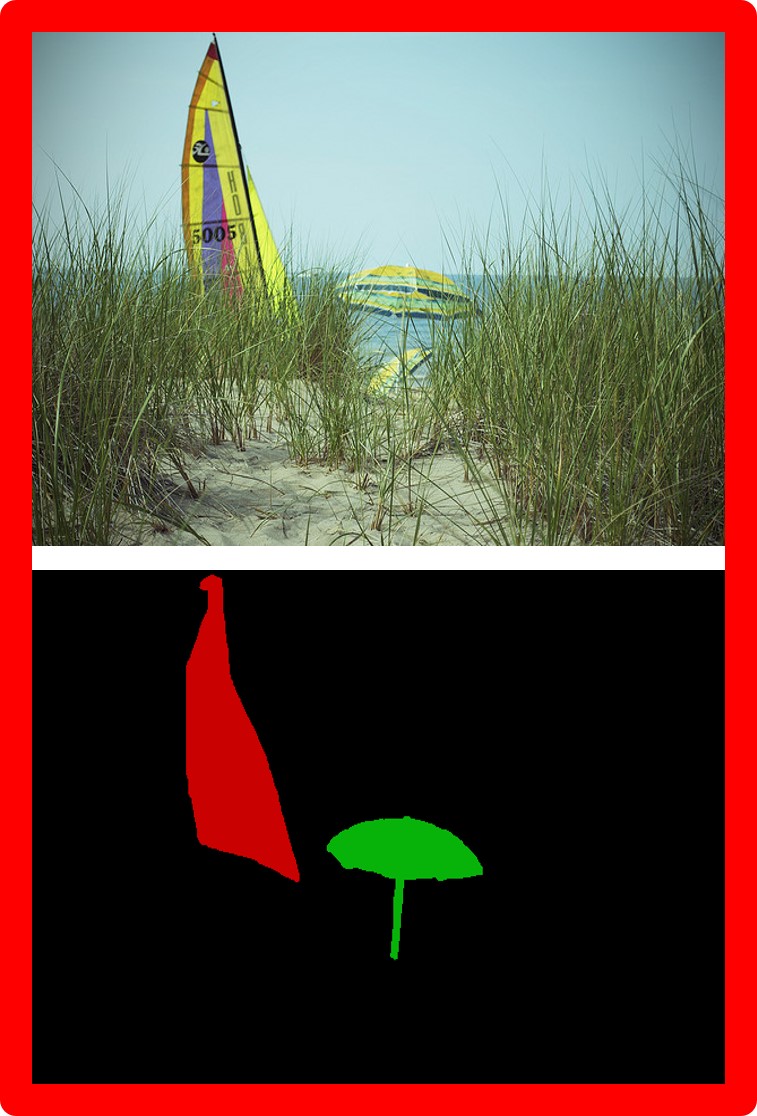} & 
	\includegraphics[width=0.118\linewidth, height=0.13\textheight]{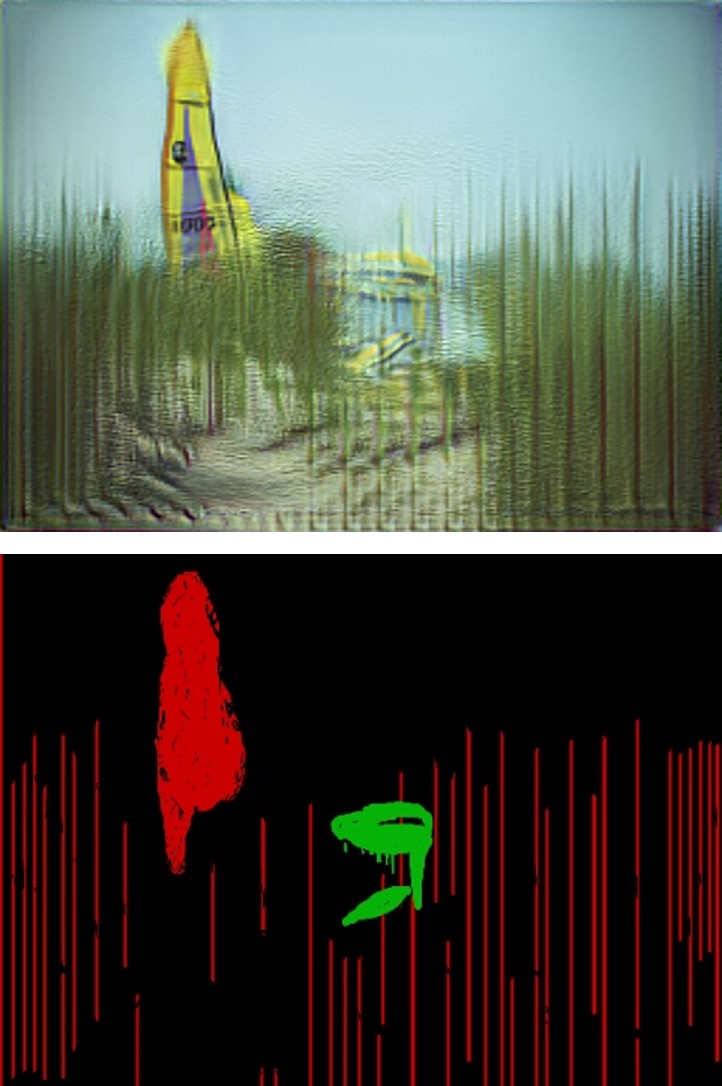} & 
	\includegraphics[width=0.118\linewidth, height=0.13\textheight]{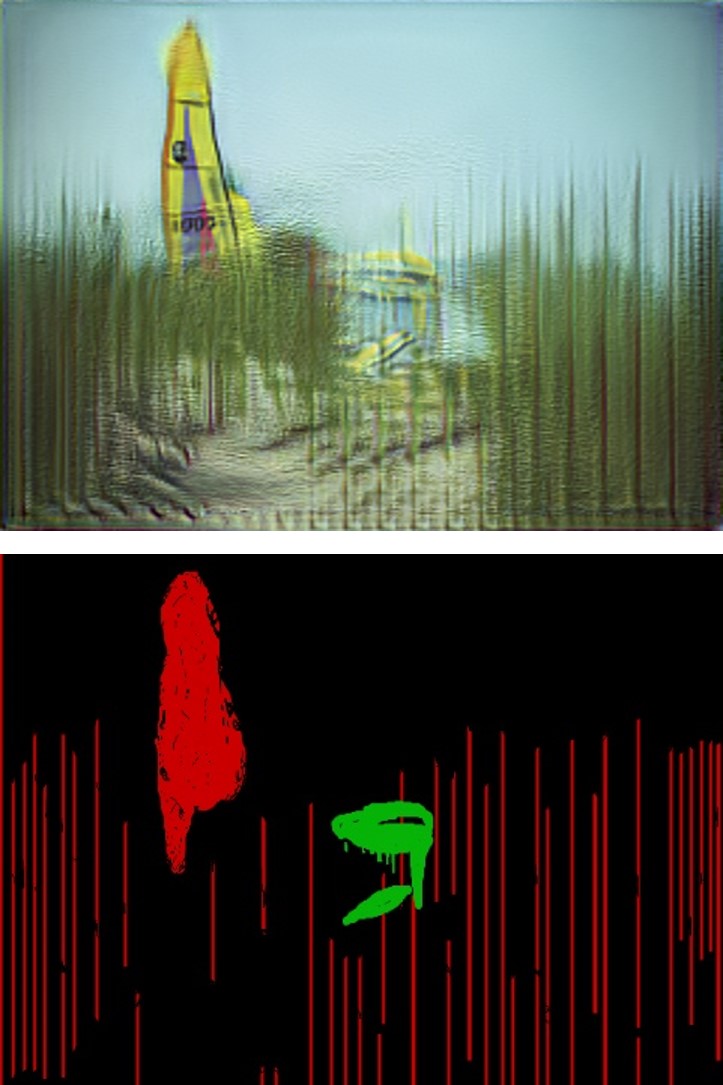} & 
	\includegraphics[width=0.118\linewidth, height=0.13\textheight]{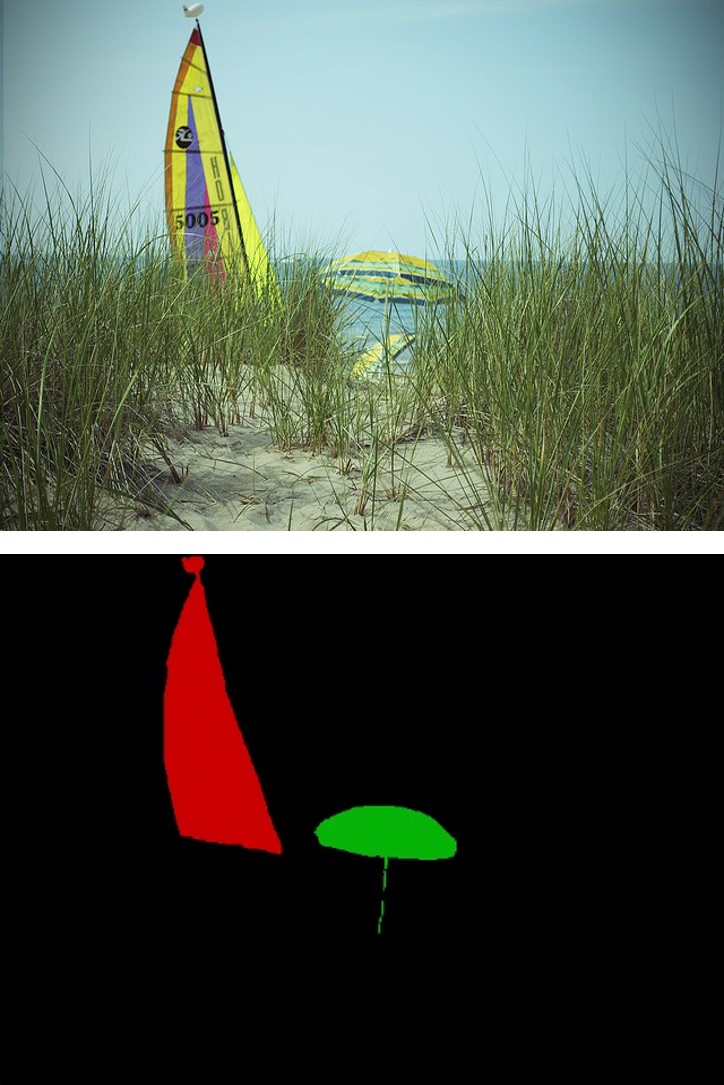}  &
	\includegraphics[width=0.118\linewidth, height=0.13\textheight]{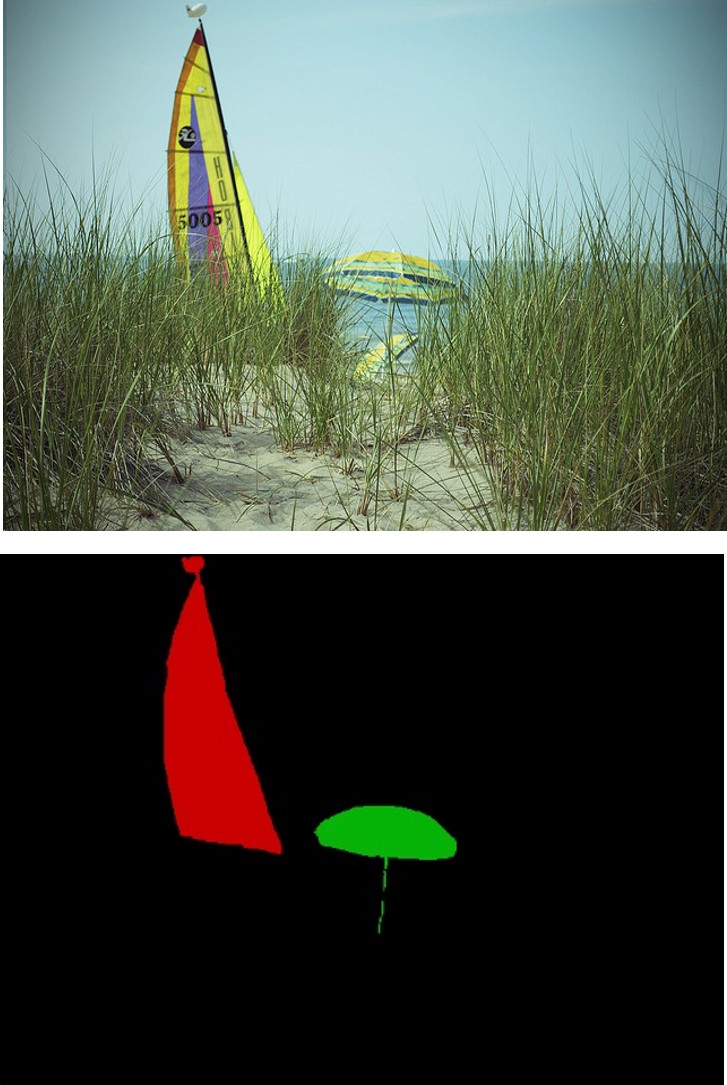} &
	\includegraphics[width=0.118\linewidth, height=0.13\textheight]{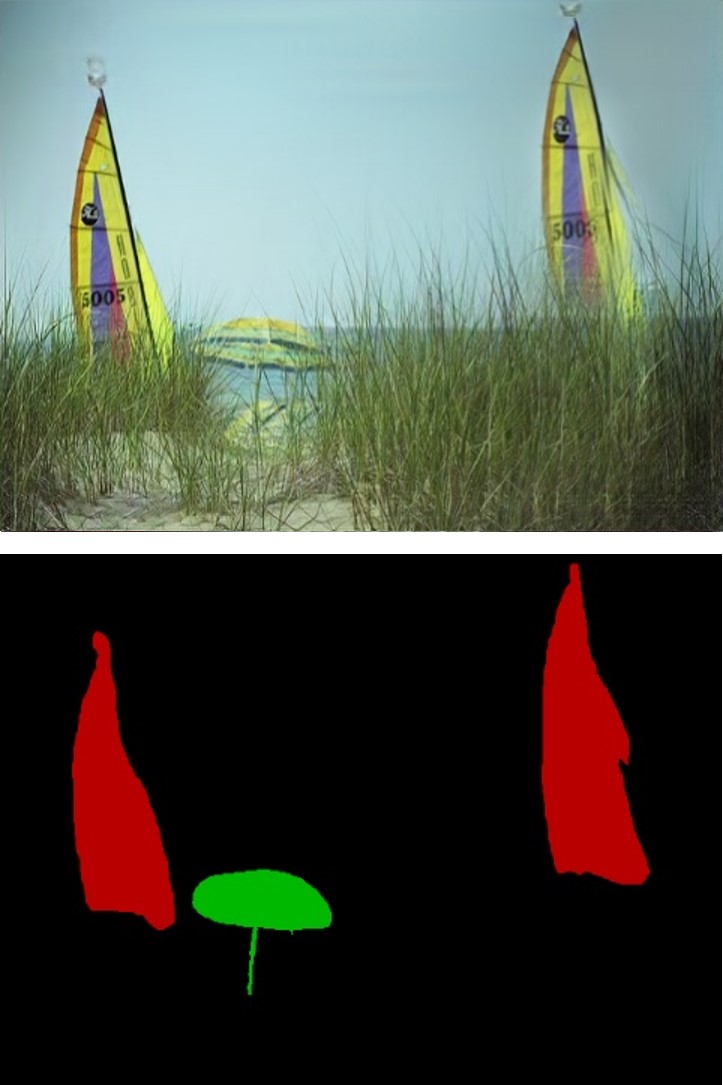} & 
	\includegraphics[width=0.118\linewidth, height=0.13\textheight]{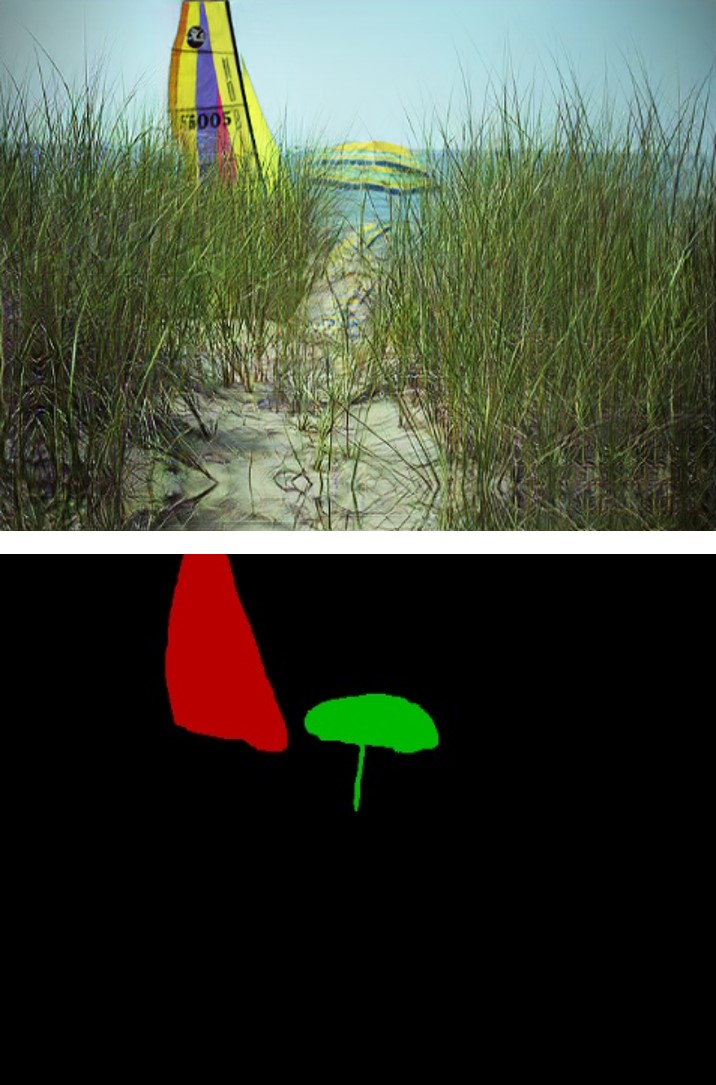} & 
	\includegraphics[width=0.118\linewidth, height=0.13\textheight]{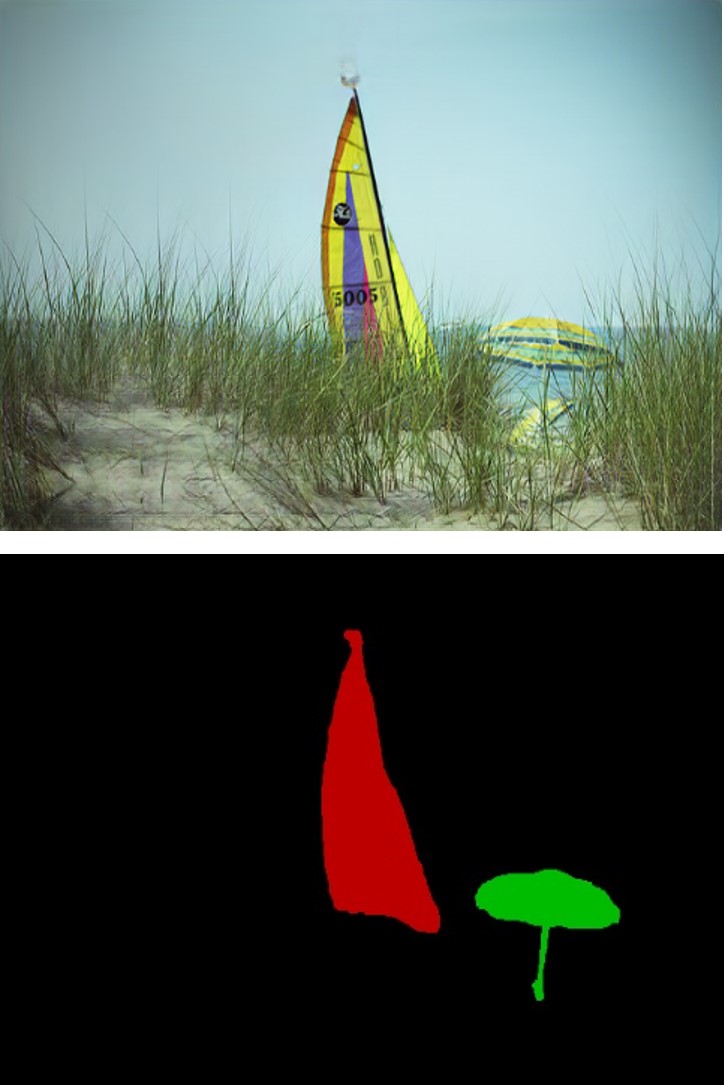}
	\tabularnewline 
	\end{tabular}
	\vspace{0.2ex}
	\caption{
A comparison to SemanticGAN \cite{li2021semantic}, trained on a single image-mask pair (in red), and DatasetGAN \cite{zhang2021datasetgan}, pre-trained on a single image and trained on a single manual mask annotation. Both models suffer from memorization, while SemanticGAN also has poor quality due to training instabilities. In contrast, OSMIS avoids mode collapse and generates diverse high-quality samples. This is achieved by means of a discriminator that judges the realism of different objects separately, which prevents memorization of the whole given image.}
	\label{fig:prior_failures}
	\vspace{-3.0ex}
\end{figure*}

%
%

Recently, several works \cite{tritrong2021repurposing, zhang2021datasetgan, li2021semantic, saha2021ganorcon} proposed to tackle this issue by jointly generating images and segmentation masks with generative adversarial networks (GANs). Utilizing a few provided pixel-level annotations in addition to an image dataset for training, such GAN models become a source of labelled data that can be used to train neural networks in various practical applications.
Despite achieving impressive synthesis of segmentation masks based on limited annotated examples, existing image-mask GAN models still require large pre-training image datasets to learn high-fidelity image synthesis. This naturally restricts their application only to the data domains where such datasets are available (e.g., images of faces or cars).
However, in some practical scenarios such a dataset can be difficult to find, for example in one-shot segmentation applications \cite{BMVC2017_167}, where the object types can be rare.
Therefore, in this work we aim to learn a high-fidelity joint mask and image synthesis
having as little limitations on the data domain as possible. To this end, we propose a novel GAN training setup, in which we assume availability only of a single training image and its segmentation mask, not relying on any image dataset for pre-training (see Fig.~\ref{fig:teaser}). After training, we aim to generate diverse new image samples and supplement them with accurate segmentation masks. To the best of our knowledge, we are the first to consider such a training scenario for GANs.



\begin{figure*}[t]
	\vspace{-1.8ex}	
	\includegraphics[width=1.0\linewidth]{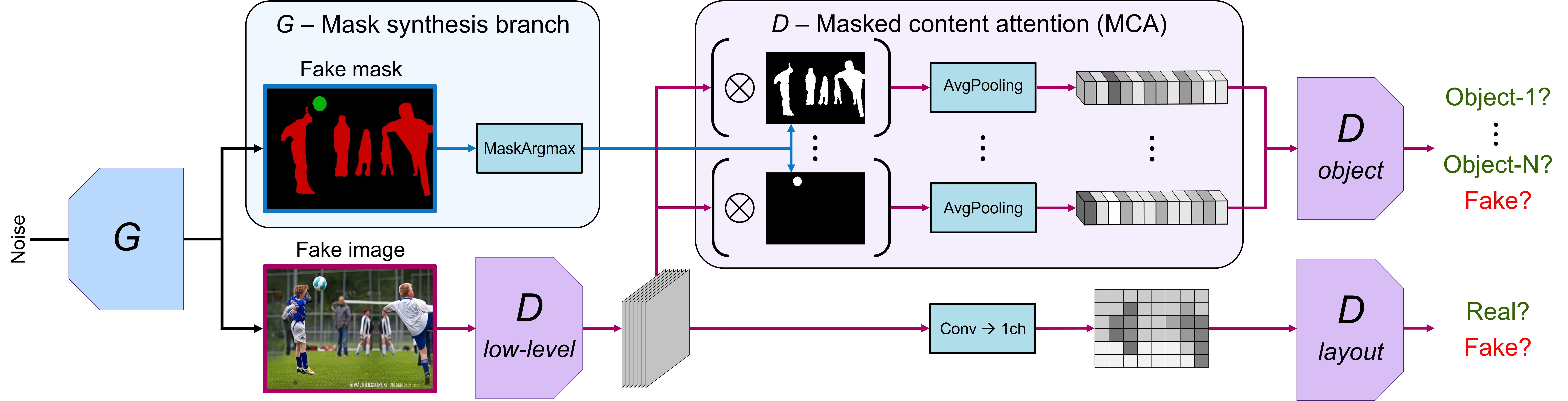}
	\vspace{-2.5ex}
	\caption{OSMIS model. A simple mask synthesis branch in the generator \textit{G} allows the generation of segmentation masks of objects together with images. The precise alignment between the masks and the generated image content is enforced by a masked content attention (MCA) module in the discriminator \textit{D}, designed to evaluate the realism of different objects separately from each other.}
	\label{fig:method}
	\vspace{-2.9ex}
\end{figure*}

Training a GAN from a single training sample is well known to be challenging due to the problem of memorization \cite{nagarajan2018theoretical}, as in many cases the generator converges to reproducing the exact copies of training data. For example, as shown in our experiments, this issue occurs in the prior image-mask GAN models from \cite{li2021semantic, zhang2021datasetgan} (see Fig.~\ref{fig:prior_failures}).
Recently, the issue of memorization has been mitigated in the line of works on single-image GANs, which enabled diverse image synthesis from a single training image \cite{Shaham2019SinGANLA, Hinz2020ImprovedTF, sushko2021one}.
Inspired by these models, we aim to extend this ability to a joint synthesis of images and segmentation masks.  
To this end, we propose a new model, introducing two modifications to conventional GAN architectures. 
Firstly, we introduce a mask synthesis branch for the generator, enabling the synthesis of segmentation masks in addition to images. Secondly, to ensure that the produced segmentation masks are precisely aligned to the generated image content, we propose a masked content attention module for the discriminator, allowing it to judge the realism of different objects separately from each other. This way, to fool the discriminator, the generator is induced to label synthesized images accurately.
In effect, 
our proposed model enables a structurally diverse, high-quality \textbf{o}ne-\textbf{s}hot joint \textbf{m}ask and \textbf{i}mage \textbf{s}ynthesis (see Fig.~\ref{fig:teaser}), and we thus name it \textbf{OSMIS}. 
As we show in our experiments, compared to prior single-image GANs \cite{Shaham2019SinGANLA, Hinz2020ImprovedTF, sushko2021one}, OSMIS not only offers an additional ability to generate accurate segmentation masks, but also achieves higher quality and diversity of generated images. 

Despite using only a single image-mask pair for training, OSMIS can generate a set of labelled samples of a high structural diversity, which sometimes cannot be achieved with standard data augmentation techniques (e.g., flipping, zooming, or rotation). For example, for a given scene, OSMIS can change the relative locations of foreground objects or edit the layout of backgrounds (see Fig.~\ref{fig:teaser}, \ref{fig:qual_results_davis}, \ref{fig:qual_results_coco}). 
Moreover, in contrast to \cite{li2021semantic, zhang2021datasetgan}, OSMIS can successfully handle masks of different types, e.g., having class-wise (see Fig.~\ref{fig:teaser}) or instance-wise (see Fig.~\ref{fig:qual_results_davis}) annotations.
This suggests a good potential of our model to serve as a source of additional labelled data augmentation for practical applications. We demonstrate this potential in Sec.~\ref{sec:exp-applications}, where we apply OSMIS at the test phase of one-shot video object segmentation \cite{pont20172017} and one-shot semantic image segmentation \cite{BMVC2017_167}. The results indicate that the data generated by OSMIS helps to improve the performance of state-of-the-art networks: OSVOS \cite{Cae+17}, STM \cite{oh2019video}, and RePRI \cite{boudiaf2021few}, providing complementary gains to standard data augmentation. We find these results promising for utilization of one-shot image-mask synthesis in future research.

\section{Related Work}
\label{sec:related_work}




\myparagraph{GANs generating segmentation masks.} 
Recently, it was observed that a GAN generator, trained on a large dataset, implicitly learns discriminative pixel-wise features of the generated scene objects \cite{tritrong2021repurposing}.
Thus, several works proposed to collect feature activations from different generator layers and transform them into a segmentation mask using a small decoder. 
RepurposeGAN \cite{tritrong2021repurposing} and DatasetGAN \cite{zhang2021datasetgan} proposed to train the decoder using a handful of manually annotated generated images. LinearGAN \cite{xu2021linear} replaced manual annotations by the predictions of an external segmentation network. 
Alternatively, SemanticGAN \cite{li2021semantic} and EditGAN \cite{ling2021editgan} enforced the alignment between generated images and masks with the loss from an additional discriminator, which takes both images and masks as inputs.

Although the above models require only a few masks to achieve high-quality image-mask synthesis, they are not successful when the number of training images is not sufficient. For example, DatasetGAN and SemanticGAN suffer from instabilities and memorization issues when trained on a single image-mask pair (see Fig.~\ref{fig:prior_failures} and \ref{supp_fig:prior_failures} in the supplementary material.).
In contrast, our model learns in this regime successfully, as it does not rely on large-scale pre-training data. As shown in experiments, this makes our model better suited for the scenarios dealing with restricted data domains, such as one-shot segmentation applications. Furthermore, our model is trained in a purely adversarial fashion without any additional overhead, e.g., not requiring manual annotations of generated images, external segmentation networks, or additional discriminators.

\myparagraph{Single Image GANs.} A line of works investigated unconditional GAN training using only a single image. Under such critically low-data regime, the models are susceptible to training instabilities, as the discriminator can simply memorize the training sample and provide uninformative gradients to the generator \cite{Karras2020TrainingGA}. SinGAN \cite{Shaham2019SinGANLA} proposed to mitigate this issue using a cascade of GANs, where each GAN stage is restricted to learn only the patch distribution at a certain image scale. 
ConSinGAN \cite{Hinz2020ImprovedTF} improved the performance and efficiency of SinGAN by re-balancing the training of different GAN stages and by training several stages concurrently. 
Since then, numerous further variations of multi-stage GAN training have been proposed \cite{arora2021singan, patch_vae_gan, bensadoun2021meta, he2021recurrent}. More recently, One-Shot GAN \cite{sushko2021one} proposed a two-branch content-layout discriminator, trained as a single stage, enabling the synthesis of images with content and layouts significantly differing from the original sample.
Our paper has a similar motivation to the above works, since we also aim to train a GAN model on a single data instance. However, we extend the single image setup with the synthesis of segmentation masks, which no prior work has considered, to the best of our knowledge.

\vspace{-0.5ex}
\section{Method}
\label{sec:method}
\vspace{-0.4ex}

Given a single image with its pixel-level segmentation mask and assuming no access to any pre-training data, we aim to generate a diverse set of new image-mask pairs. In this section, we present OSMIS, our one-shot image-mask synthesis model. Adopting One-Shot GAN \cite{sushko2021one} as a state-of-the-art image synthesis baseline (Sec.~\ref{sec:meth-SIV-GAN}), we propose modifications to the generator and discriminator architecture, enabling one-shot synthesis of segmentation masks that are precisely aligned with generated images (Sec.~\ref{sec:meth-MCA}). 




\subsection{One-Shot GAN baseline}
\label{sec:meth-SIV-GAN}


As the baseline network architecture, we select the state-of-the-art model One-Shot GAN \cite{sushko2021one}, as
it achieves the highest quality and diversity of one-shot image synthesis among previous works.
One-Shot GAN proposed a two-branch discriminator, in which an input image $x$ is first transformed into a feature representation $F(x)$ by a low-level discriminator $\mathcal{D}_{low-level}$. Next, two separate discriminators assess different aspects of $F(x)$. The content discriminator $\mathcal{D}_{content}$ judges the realism of objects regardless of their spatial location by averaging out the spatial information contained in $F(x)$ via global average pooling. On the other hand, the layout discriminator $\mathcal{D}_{layout}$ evaluates the realism only of the spatial scene layouts by squeezing $F(x)$ with a one-channel convolution. In addition, the discriminator applies feature augmentation in the content and layout representations of $F(x)$ to further increase the high-level diversity among generated samples. The adversarial loss of the One-Shot GAN model consists of three terms:

\vspace{-2ex}
\begin{equation} 
\mathcal{L}_{adv} (G,D)  = \mathcal{L}_{D_{content}} + \mathcal{L}_{D_{layout}} + 2 \mathcal{L}_{D_{low\text{-}level}}, 
\label{eq:loss_oneshotgan}
\end{equation}
where each term is the mean of binary cross entropies obtained at different layers of respective discriminator parts. 


\subsection{OSMIS model}
\label{sec:meth-MCA}


In contrast to one-shot image synthesis, we assume that the single training image is provided with its pixel-level mask of objects, not assuming any fixed annotation type (e.g., class-wise or instance-wise). To incorporate it into the training process, we introduce two modifications to the architecture of the baseline model.
Firstly,
we propose to generate segmentation masks simultaneously with 
images via an additional generator's mask synthesis branch. 
Secondly, to enforce the precise mask alignment to the generated image content, we re-formulate the objective of the content discriminator $\mathcal{D}_{content}$, designing it to judge the fidelity of different objects separately from each other. This is made possible by the introduced masked content attention module, which builds a separate content feature vector for each object considering the provided segmentation mask. The overview of our model architecture is shown in Fig.~\ref{fig:method}. Next, we describe the proposed modifications in detail.

\myparagraph{Mask synthesis branch in the generator.} 
In line with \cite{tritrong2021repurposing, zhang2021datasetgan}, we hypothesize that during training the generator should be able to learn discriminative features that completely describe the appearance 
of generated objects. 
Thus, while synthesizing an image, we collect feature activations of the generator layers and use them as input for the mask synthesis branch. 
In contrast to \cite{tritrong2021repurposing, zhang2021datasetgan}, 
we use only the activations after the last generator block, as this simplest solution already performs well in our experiments. Using a simple convolution followed by a softmax activation, we transform these features into an $N$-channel soft probability map, where each channel corresponds to one of $N-1$ objects of interest in the segmentation mask or to the background. To obtain the final discrete mask prediction, an argmax operation $T$ along the channel dimension is applied.

To enable the training of the mask synthesis branch with the discriminator loss, the generated masks should allow back-propagation 
of gradients, similarly to generated images. In our experiments, feeding the discriminator the continuous segmentation probability maps obtained before the non-differentiable argmax operation $T$ impaired the GAN training, as the discriminator learnt to detect the continuous-discrete discrepancy between fake and real inputs. Thus, inspired from \cite{NIPS2017_7a98af17, bengio2013estimating}, we enable back-propagation through argmax by developing a straight-through gradient estimator:
\vspace{-0.5ex}
\begin{equation} 
\mathrm{MaskArgmax}(y) = y + T(y) - sg[y],
\end{equation}
\vspace{-0.5ex}
where $sg$ denotes a stop-gradient operation. This way, the discriminator is provided with the generated masks in a discrete form $T(y)$, which enables its effective training, while the generator can be trained with the gradients passing through its probability map prediction $y$.

Yet, this solution can sometimes lead to degenerate solutions, e.g., when all the pixels are predicted as the background channel. This cannot be corrected during training, as in this case the gradient flow through all the other mask channels is blocked. We found that it can be mitigated by softening the argmax operation $T$ at the beginning of training. For this, during the first $P_0$ epochs we regard each mask pixel as a random variable following Bernoulli distribution:

\vspace{-0.5ex}
\begin{equation} 
T(y) =
\begin{cases}
\sim \mathrm{Bernoulli}(y)  &\text{epoch} < P_0, \\
\mathrm{argmax}(y) &\text{epoch} \geq P_0.
\end{cases}
\end{equation}
\vspace{-0.5ex}

\myparagraph{Masked content attention in the discriminator.} 
To provide a training signal to the generator's mask synthesis branch, we propose to incorporate the learning of the image-mask alignment to the objective of the content discriminator $\mathcal{D}_{content}$. In \cite{sushko2021one},  $\mathcal{D}_{content}$ was designed to judge the content distribution of the whole given image. Considering the provided segmentation mask, we can now select the image areas belonging to different objects,
and require the discriminator to learn their appearance separately from each other. 
With this objective, as the discriminator can compare the appearance of the area belonging to the same object in real and fake images, it encourages the generator not only to synthesize realistic objects, but also to label them correctly.

To this end, we introduce a masked content attention ($\mathrm{MCA}$) module. As shown in Fig.~\ref{fig:method}, $\mathrm{MCA}$ receives a downsampled segmentation mask $y$ along with an intermediate feature representation $F(x)=\mathcal{D}_{low-level}(x)$ of an input image $x$, and thereout produces
a set of $N$ content vectors, corresponding to the masked content representations of each of the $N-1$ objects of interest and the background:

\begin{figure*}[t]
\begin{centering}
\setlength{\tabcolsep}{0.0em}
\par\end{centering}
\renewcommand{\arraystretch}{1.25}
\vspace{-1.2ex}
\begin{tabular}{@{\hskip 0.02in}c@{\hskip 0.06in}c@{\hskip 0.04in}c@{\hskip 0.04in}c@{\hskip 0.06in}|@{\hskip 0.06in}c@{\hskip 0.06in}c@{\hskip 0.04in}c@{\hskip 0.04in}c@{\hskip 0.04in}c}

\small Training pair & \multicolumn{3}{c}{\small Generated samples} & \small Training pair & \multicolumn{3}{c}{\small Generated samples} 
\tabularnewline 

\includegraphics[width=0.117\linewidth, height=0.135\textheight]{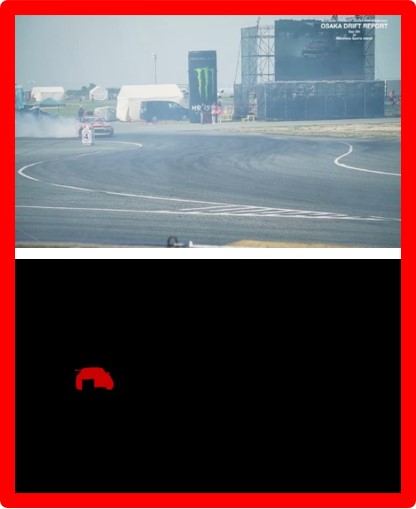} & 
\includegraphics[width=0.117\linewidth, height=0.135\textheight]{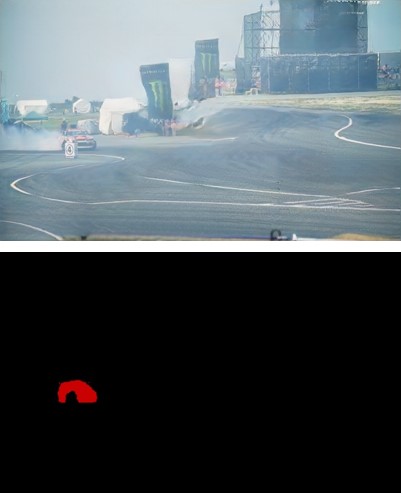} & 
\includegraphics[width=0.117\linewidth, height=0.135\textheight]{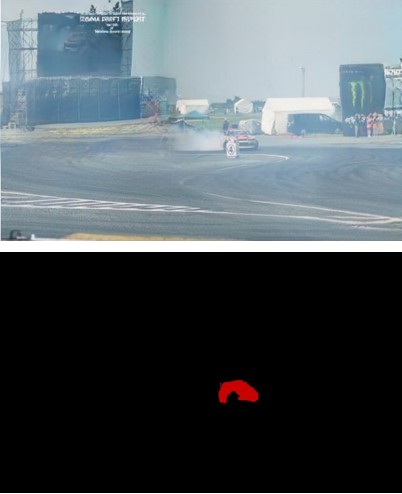} & 
\includegraphics[width=0.117\linewidth, height=0.135\textheight]{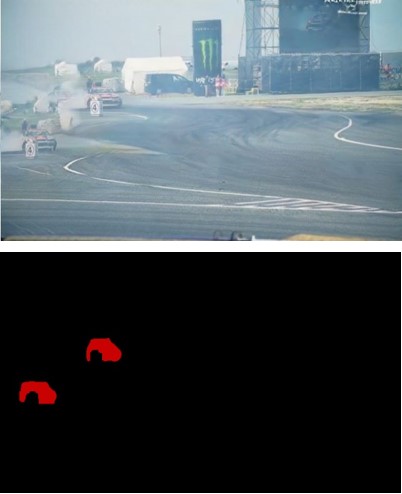}  &

\includegraphics[width=0.117\linewidth, height=0.135\textheight]{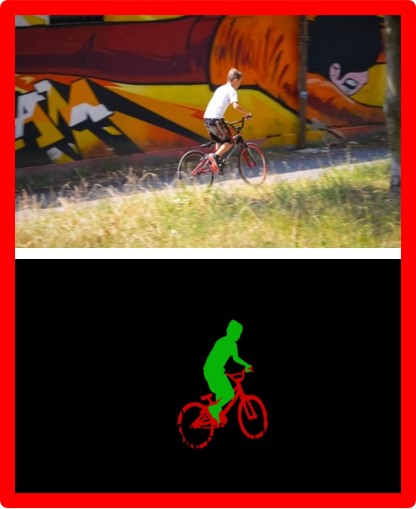} & 
\includegraphics[width=0.117\linewidth, height=0.135\textheight]{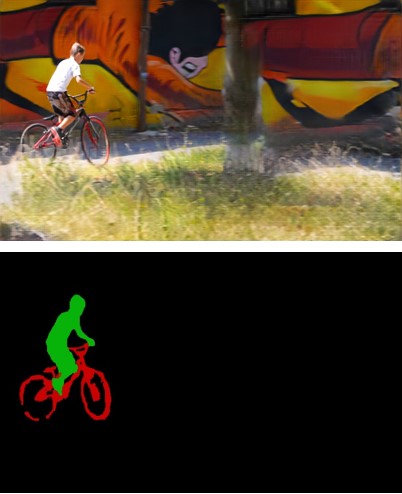} & 
\includegraphics[width=0.117\linewidth, height=0.135\textheight]{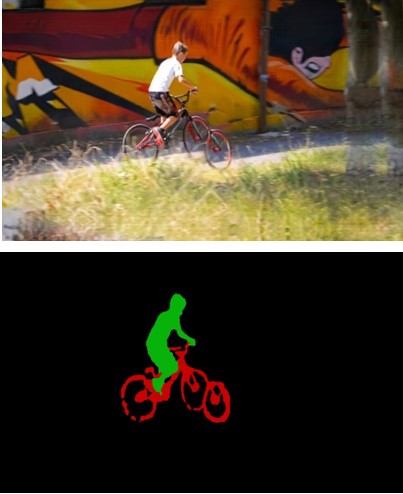} & 
\includegraphics[width=0.117\linewidth, height=0.135\textheight]{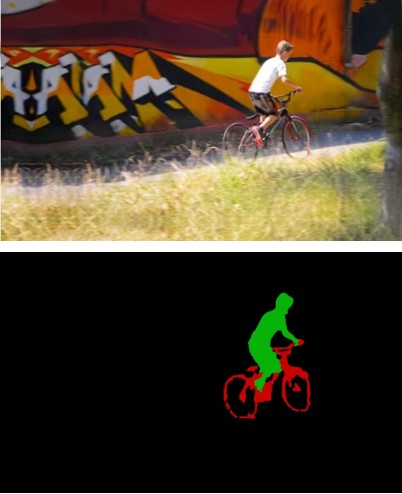}  
\tabularnewline 

\includegraphics[width=0.117\linewidth, height=0.135\textheight]{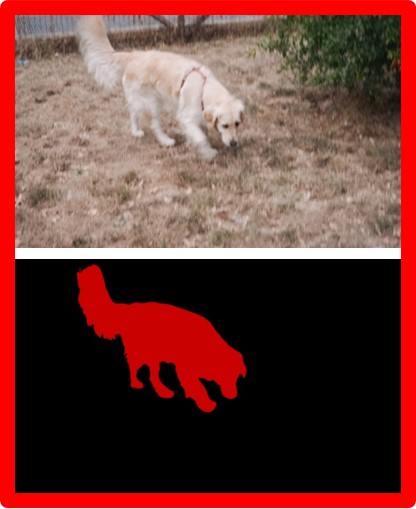} & 
\includegraphics[width=0.117\linewidth, height=0.135\textheight]{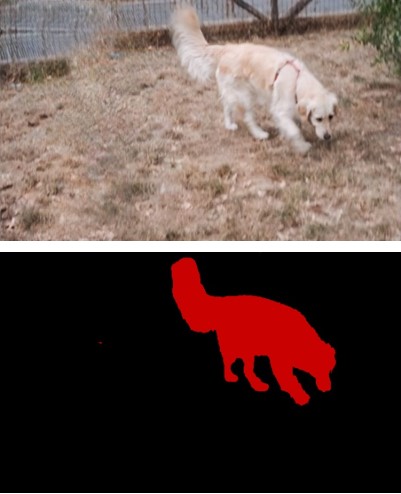} & 
\includegraphics[width=0.117\linewidth, height=0.135\textheight]{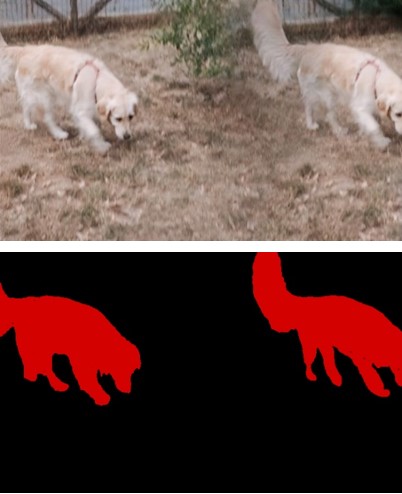} & 
\includegraphics[width=0.117\linewidth, height=0.135\textheight]{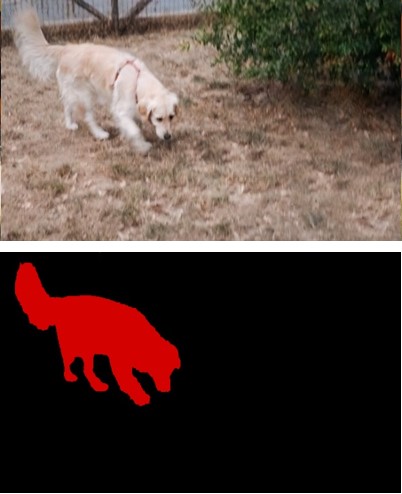} & 

\includegraphics[width=0.117\linewidth, height=0.135\textheight]{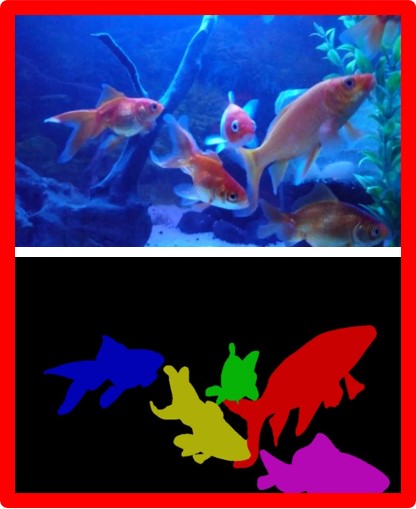} & 
\includegraphics[width=0.117\linewidth, height=0.135\textheight]{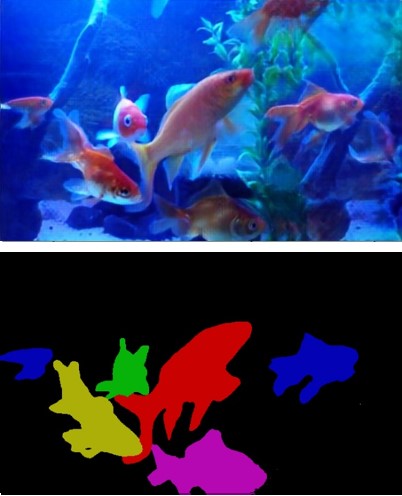} & 
\includegraphics[width=0.117\linewidth, height=0.135\textheight]{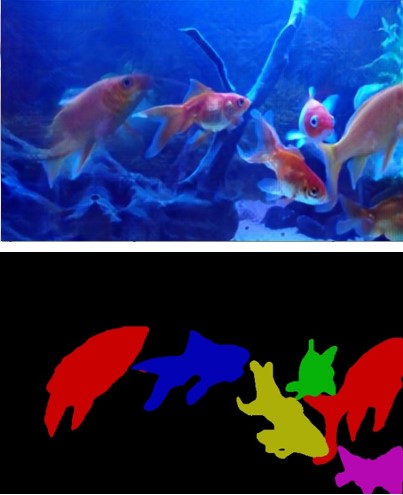} & 
\includegraphics[width=0.117\linewidth, height=0.135\textheight]{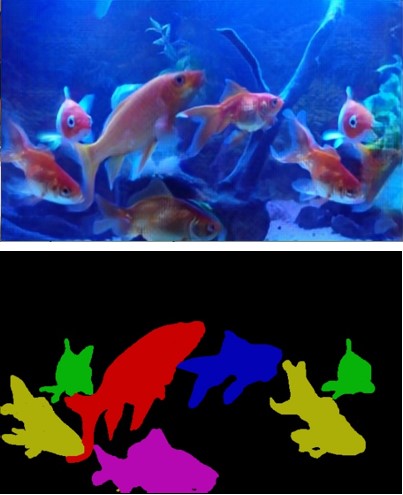} 
\tabularnewline
\end{tabular}
\vspace{-0.5ex}
\caption{Qualitative results of OSMIS on DAVIS \cite{pont20172017}. Given a single image-mask pair for training, our model achieves high-fidelity image synthesis with a high structural diversity, changing the positions of objects or editing the layout of backgrounds. For each synthesized image, it produces segmentation masks that accurately annotate the generated content. Training pairs are shown in red frames.}
\label{fig:qual_results_davis}
\vspace{-1.0ex}
\end{figure*}

\begin{figure*}[t]
\begin{centering}
\setlength{\tabcolsep}{0.0em}
\par\end{centering}
\renewcommand{\arraystretch}{1.25}
\begin{tabular}{@{\hskip 0.02in}c@{\hskip 0.06in}c@{\hskip 0.04in}c@{\hskip 0.04in}c@{\hskip 0.06in}|@{\hskip 0.06in}c@{\hskip 0.06in}c@{\hskip 0.04in}c@{\hskip 0.04in}c@{\hskip 0.04in}c}

\small Training pair & \multicolumn{3}{c}{\small Generated samples} & \small Training pair & \multicolumn{3}{c}{\small Generated samples} 
\tabularnewline 

\includegraphics[width=0.117\linewidth, height=0.135\textheight]{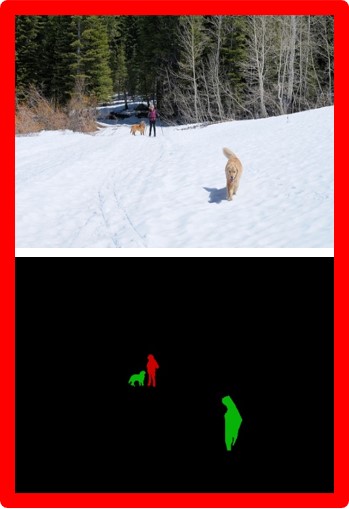} & 
\includegraphics[width=0.117\linewidth, height=0.135\textheight]{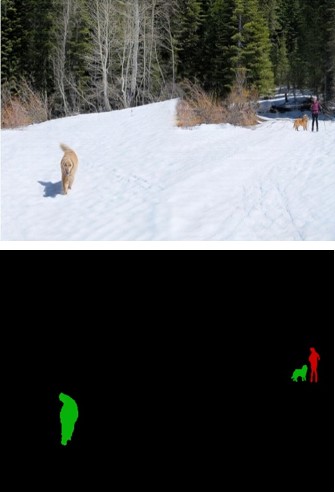} & 
\includegraphics[width=0.117\linewidth, height=0.135\textheight]{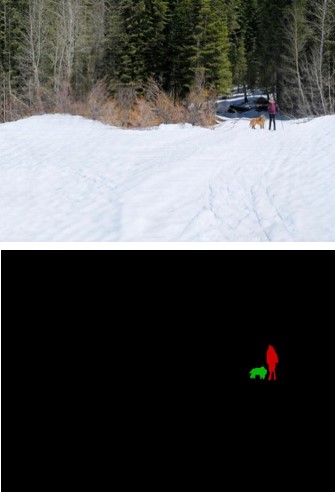} & 
\includegraphics[width=0.117\linewidth, height=0.135\textheight]{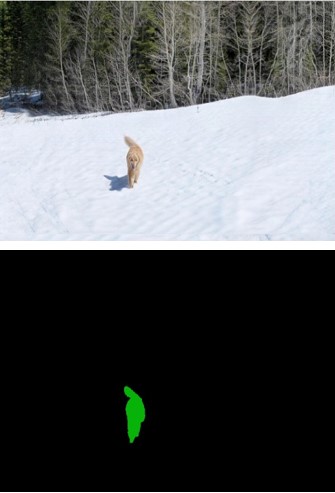} &

\includegraphics[width=0.117\linewidth, height=0.135\textheight]{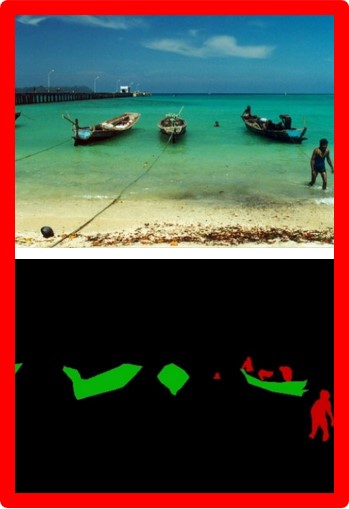} & 
\includegraphics[width=0.117\linewidth, height=0.135\textheight]{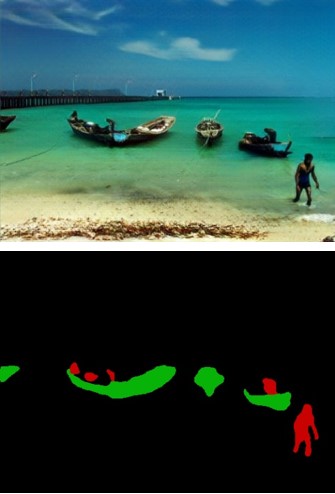} & 
\includegraphics[width=0.117\linewidth, height=0.135\textheight]{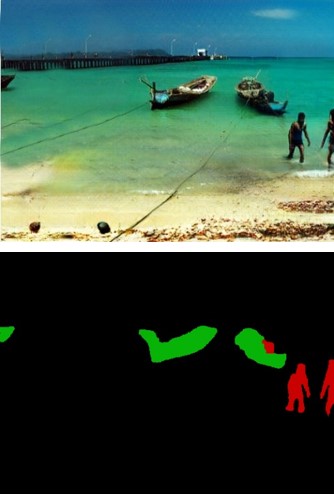} & 
\includegraphics[width=0.117\linewidth, height=0.135\textheight]{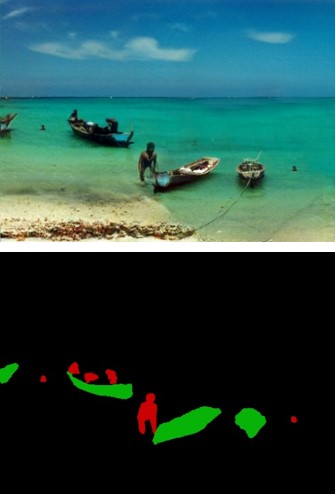} 
\tabularnewline

\includegraphics[width=0.117\linewidth, height=0.135\textheight]{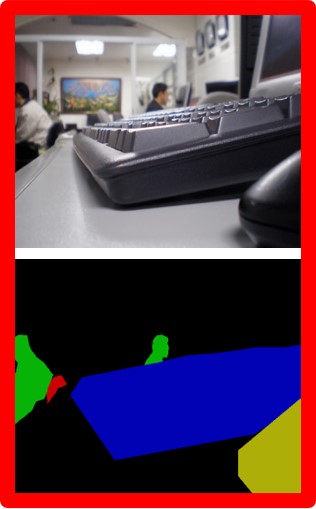} & 
\includegraphics[width=0.117\linewidth, height=0.135\textheight]{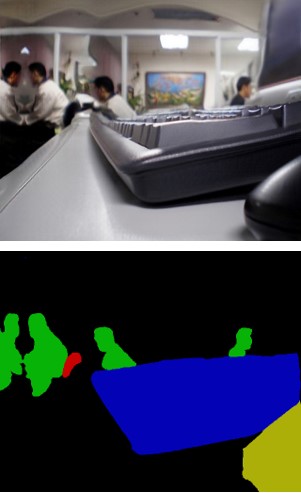} & 
\includegraphics[width=0.117\linewidth, height=0.135\textheight]{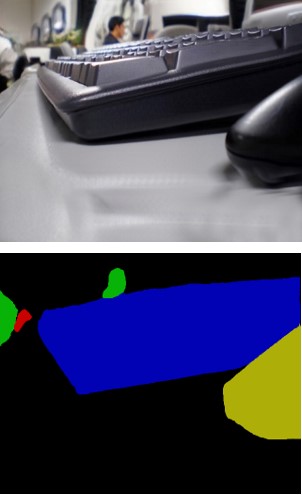} & 
\includegraphics[width=0.117\linewidth, height=0.135\textheight]{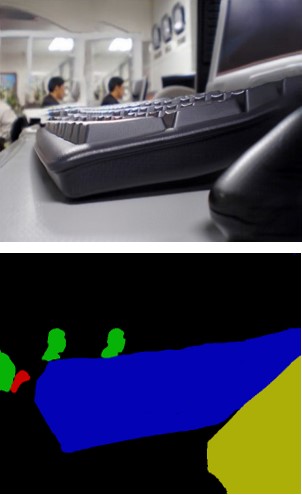} &

\includegraphics[width=0.117\linewidth, height=0.135\textheight]{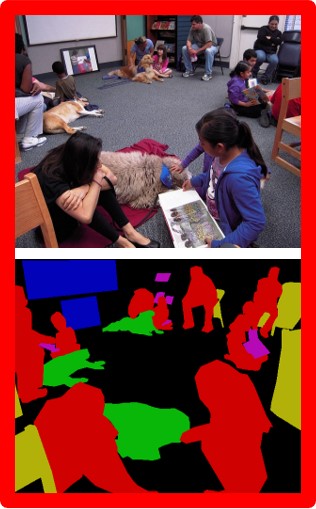} & 
\includegraphics[width=0.117\linewidth, height=0.135\textheight]{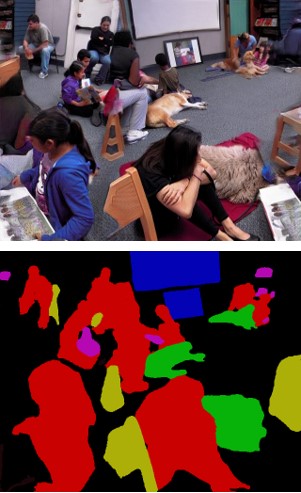} & 
\includegraphics[width=0.117\linewidth, height=0.135\textheight]{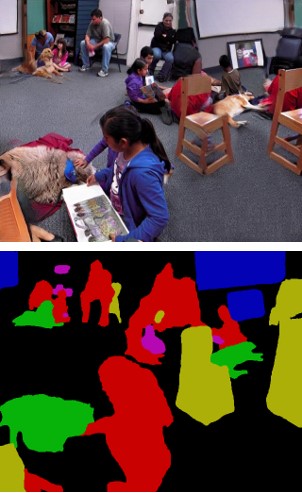} & 
\includegraphics[width=0.117\linewidth, height=0.135\textheight]{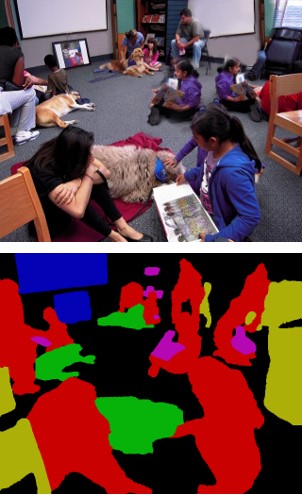} 
\tabularnewline

\end{tabular}
\vspace{-0.5ex}
\caption{Qualitative results of OSMIS on COCO \cite{Lin2014MicrosoftCC}. OSMIS successfully deals with different scene types and annotation styles. For example, it achieves high quality and diversity for both indoor and outdoor scenes, or sparse and dense annotations of foreground objects.}
\label{fig:qual_results_coco}
\vspace{-1.0ex}
\end{figure*}

\vspace{-2.5ex}
\begin{equation}
\mathrm{MCA}(x, y) = \{\mathrm{AvgPool} \left(F(x) \times \mathbbm{1}_{y=i} \right) \}_{i=1}^{N}.
\label{eq:MCA}
\end{equation}
\vspace{-2.5ex}


Accordingly, we re-design the objective of the content discriminator (further denoted $\mathcal{D}_{object}$). 
For each of the obtained object representations, our proposed $\mathcal{D}_{object}$ is induced to predict a correct identity 
of each object or background of a real image, while all the identities of fake images should be categorized as an additional fake class:
\vspace{-0.5ex}
\begin{equation} 
\begin{aligned}
\mathcal{L}_{D_{object}} = &-\mathbbm{E}_{(x, y)}\left[\sum\limits_{i=1}^{N} \alpha_{i} \log \mathcal{D}_{object}^i(\mathrm{MCA}^{i}(x,y))\right] \\
&- \mathbbm{E}_z\left[\sum\limits_{i=1}^{N} \log (1-\mathcal{D}_{object}^{fake}(\mathrm{MCA}^i(G(z)))\right],
\end{aligned}
\label{eq:loss_content}
\end{equation}
\vspace{-0.5ex}
where $z$ is the noise vector used by the generator $G$ to synthesize a fake image-mask pair $G(z) = \{G_x(z), G_y(z)\}$, $(x,y)$ denotes the real image-mask pair, and $\mathcal{D}^i(*)$ is the discriminator logit for the object $i$.
Considering that different 
objects or background can occupy different areas, we introduce a class balancing weight $\alpha_i$, which is the inverse of the per-pixel class frequency in the segmentation mask $y$:
\vspace{-0.5ex}
\begin{equation} 
\alpha_i = \frac{(\mathrm{sum}(\mathbbm{1}_{y=i}))^{-1}}{\sum_{j=1}^N (\mathrm{sum}(\mathbbm{1}_{y=j}))^{-1}}.
\label{eq:alpha}
\end{equation}
Note that the balancing is applied only for real images, as in Eq.~\ref{eq:loss_content} all fake objects are considered as the same class.


Our $\mathcal{D}_{object}$ 
learns the content distribution of each 
object separately. The advantage of such a training scheme is two-fold. Firstly, a generator now needs to synthesize correct segmentation masks in order to fool the discriminator. The precise image-mask alignment is thus enforced directly by the adversarial loss, without the need for using additional networks or manual annotation. Secondly, as $\mathrm{MCA}$ provides representations only of separate objects, $\mathcal{D}_{object}$ has restricted access to the content distribution of the whole image. In effect, the discriminator memorization of the whole training sample becomes more difficult, which enables more diverse image synthesis (see Table \ref{table:comparison}).

\section{Experiments}
\label{sec:experiments}

We evaluate our model as follows. Firstly, we provide the qualitative and quantitative assessment of the achieved one-shot image-mask synthesis, evaluating the quality and diversity of generated images, as well as their alignment to the produced segmentation masks (Sec.~\ref{sec:exp-synthesis}).
Secondly, we apply OSMIS to two one-shot segmentation applications, demonstrating the potential of the generated image-mask pairs to be used as data augmentation (Sec.~\ref{sec:exp-applications}).

\subsection{Evaluation of one-shot image-mask synthesis}
\label{sec:exp-synthesis}


\myparagraph{Training details.} We train our model with the loss from Eq.~\eqref{eq:loss_content} for the object discriminator $\mathcal{D}_{object}$, setting $P_0$=$15000$. We employ differentiable augmentation (DA) of input images and masks while training the discriminator, using the whole set of transformations as proposed in \cite{Karras2020TrainingGA}. We use an exponential moving average of the generator weights with a decay of 0.9999, and follow \cite{sushko2021one} in setting all the other hyperparameters.
More training details are shown in the supplementary material.

\myparagraph{Datasets.} To evaluate the synthesis, we use the DAVIS dataset \cite{pont20172017}, originally introduced for video object segmentation. For each video from the DAVIS-17 validation split, we take the first frame and its segmentation mask of objects, which results in 30 image-mask pairs on which we train separate models. The resolution is set to 640x384. For additional visual results, we use samples from COCO \cite{Lin2014MicrosoftCC}, trying to closely fit their resolution. Note that the datasets have different annotation types (class-wise and instance-wise).

\begin{table}[t]

\centering
\vspace{-0.5ex}
\begin{tabular}{l|c|c}

		\multirow{1}{*}{Method} & \multirow{1}{*}{SIFID$\downarrow$}  & \multirow{1}{*}{LPIPS$\uparrow$} 
		\tabularnewline 
		
		\hline 	
		
		SinGAN \cite{Shaham2019SinGANLA} & 0.131 & 0.267 
		\tabularnewline
		ConSinGAN \cite{Hinz2020ImprovedTF} & 0.103 & 0.296 
		\tabularnewline 
		One-Shot GAN \cite{sushko2021learning} & 0.091 & 0.347 
		\tabularnewline 		
		OSMIS (ours) & \textbf{0.073} & \textbf{0.387} 
\end{tabular}
\vspace{1.4ex}
\caption{Comparison of image quality and diversity to single-image GANs on DAVIS-17. Bold denotes the best performance.}
\label{table:ablations} %
\vspace{-0.0ex}
\end{table}

\begin{table}[t]

\centering
\begin{tabular}{l|c|c|c}

		\multirow{1}{*}{Method} & \multirow{1}{*}{SIFID$\downarrow$}  &  \multirow{1}{*}{LPIPS$\uparrow$}  & \multirow{1}{*}{mIoU} 
		\tabularnewline 
		
		\hline 	
		
		DatasetGAN \cite{zhang2021datasetgan} & 0.118 & \color{darkred}0.007\color{black}  & ~\color{darkgray}91.1*\color{black}
		\tabularnewline
		SemanticGAN \cite{li2021semantic} & 0.211 & \color{darkred}0.012\color{black}  & 65.8
		\tabularnewline 
		OSMIS (ours) & \textbf{0.073} & \textbf{0.387}  & \textbf{86.6} 
\end{tabular}
\vspace{1.4ex}
\caption{Comparison to prior image-mask GANs on DAVIS-17. Bold denotes the best performance. Red indicates mode collapse. * Indicates manual annotation of masks for DatasetGAN training.}
\label{table:ablation_imagemask} %
\vspace{-2.0ex}
\end{table}

\myparagraph{Metrics.} To mind a possible quality-diversity trade-off in our one-shot regime \cite{robb2021fewshot, li2020few}, we assess the quality and diversity of generated images separately. For this, we report the average SIFID \cite{Shaham2019SinGANLA} as the measure of image quality, while the average LPIPS \cite{zhang2018unreasonable} between the pairs of generated images is used to assess the diversity of synthesis.

On the other hand, evaluating the quality of generated masks is challenging, because generated images do not have ground truth segmentation annotations. To bypass this issue, we propose to evaluate the alignment between generated masks and synthetic images using an external segmentation network. For this, we take a UNet \cite{Ronneberger2015UNetCN} and train it on the generated image-mask pairs for 500 epochs. After training, we compute its mIoU performance on the original real image, augmented with standard geometric transformations. Intuitively, a good performance on this test reveals that synthetic masks describe well the objects from the real data, indicating precise alignment between the generated images and their masks.

\myparagraph{Qualitative results.} 
Fig.~\ref{fig:qual_results_davis} and~\ref{fig:qual_results_coco} show image-mask pairs generated by OSMIS trained on samples from DAVIS and COCO. Given only a single image-mask pair, our model learns to generate new image-mask pairs, demonstrating a remarkable structural diversity among samples, photorealism of synthesized images, and a high quality of generated annotations. For example, OSMIS can re-synthesize the provided scene with a different number of foreground objects, e.g., more dogs (3$^\text{rd}$ example in Fig.~\ref{fig:qual_results_davis}), less people (2$^\text{nd}$ example in Fig.~\ref{fig:qual_results_coco}), or edit layouts of backgrounds (1$^\text{st}$ examples in Fig.~\ref{fig:qual_results_davis}-\ref{fig:qual_results_coco}), in all cases providing accurate segmentation masks for the re-synthesized scenes.
We note that reaching such structural differences to training data simultaneously with photorealism is extremely difficult from a single sample. For example, 
it could not be achieved with DatasetGAN or SemanticGAN due to memorization issues and training instabilities (see Fig.~\ref{fig:prior_failures}).
Lastly, we remark that OSMIS successfully deals with very different scene types (e.g., both indoor and outdoor scenes), supports masks with both sparse and dense object annotations (e.g., foreground objects occupying small or large image areas), and can handle masks with many objects or even separate instances of the same semantic class (e.g., fish in 4$^\text{th}$ example in Fig.~\ref{fig:qual_results_davis}).

\myparagraph{Quantitative results.} We compare the quality and diversity of generated \textit{images} to the single-image GAN models SinGAN \cite{Shaham2019SinGANLA}, ConSinGAN \cite{Hinz2020ImprovedTF} and One-Shot GAN \cite{sushko2021one}. The image-mask synthesis is compared to the previous methods DatasetGAN \cite{zhang2021datasetgan} and SemanticGAN \cite{li2021semantic}. 
We use the official repositories provided by the authors.


\begin{table}[t]
\centering
\vspace{-0.5ex}
\begin{tabular}{@{\hskip 0.03in}l@{\hskip 0.05in}|@{\hskip 0.08in}c@{\hskip 0.08in}|@{\hskip 0.08in}c@{\hskip 0.08in}|@{\hskip 0.08in}c@{\hskip 0.03in}}
	
	\multirow{1}{*}{Mask supervision} & \multirow{1}{*}{SIFID$\downarrow$}  & \multirow{1}{*}{LPIPS$\uparrow$} & \multirow{1}{*}{mIoU} 
	
	\tabularnewline	
	\hline 	
	
	None & 0.071 & 0.368  & -  
	\tabularnewline 
	\arrayrulecolor{lightgray}	\hline \arrayrulecolor{black}
	Projection \cite{miyato2018cgans} & \textbf{0.071} & \color{darkred} 0.362 &  72.1 
	\tabularnewline
	Input concat. & 0.079 & \color{darkred} 0.328  & 82.4 
	\tabularnewline
	SemanticGAN  $D_m$ \cite{li2021semantic} & 0.074 & \color{darkred} 0.351 & 83.3 
	\tabularnewline
	MCA (ours) & 0.073 & \textbf{0.387} & \textbf{86.6} 
\end{tabular}
\vspace{0.9ex}
\caption{Comparison of MCA to other mask synthesis supervision mechanisms on DAVIS-17. 
	Red indicates decreased diversity compared to the baseline. Bold denotes the best performance.}
\label{table:comparison} %
\vspace{-2.6ex}
\end{table}

\begin{figure*}[t]
	\begin{centering}
		\setlength{\tabcolsep}{0.0em}
		\par\end{centering}
	\renewcommand{\arraystretch}{1.00}
	\centering
	\vspace{-0.5ex}
	\begin{tabular}{@{\hskip -0.04in}c@{\hskip 0.03in}c@{\hskip 0.03in}c@{\hskip 0.03in}c@{\hskip 0.03in}c@{\hskip 0.03in}c}
		
		\small Training pair & $N_{\text{low-level}}=1$ & $N_{\text{low-level}}=2$ & $N_{\text{low-level}}=3$ & $N_{\text{low-level}}=4$ & $N_{\text{low-level}}=5$ 

		\tabularnewline 
		
		\multirow{-9}{*}{ \begin{tabular}{@{\hskip 0.00in}c@{\hskip 0.00in}} \vspace{-1.5ex} \includegraphics[width=0.163\linewidth]{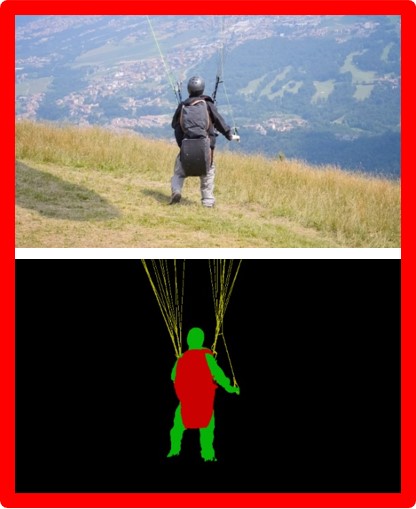} \end{tabular}} &

		\includegraphics[width=0.16\linewidth]{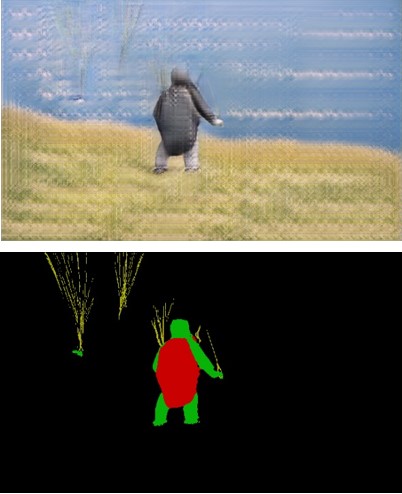} & 
		\includegraphics[width=0.16\linewidth]{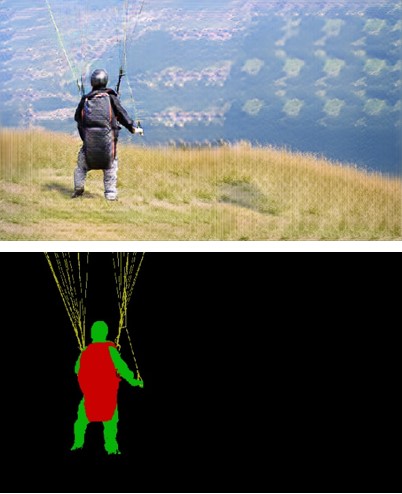} & 
		\includegraphics[width=0.16\linewidth]{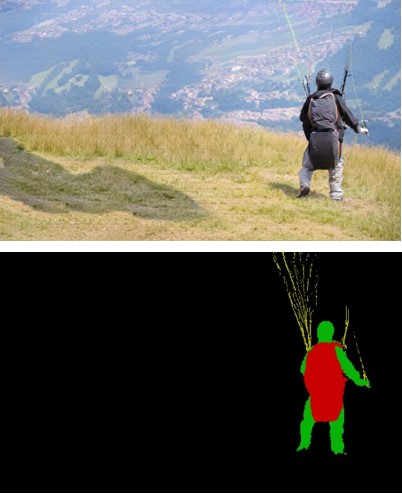} & 
		\includegraphics[width=0.16\linewidth]{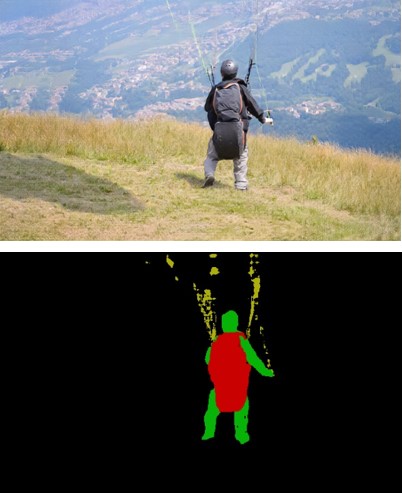} & 
		\includegraphics[width=0.16\linewidth]{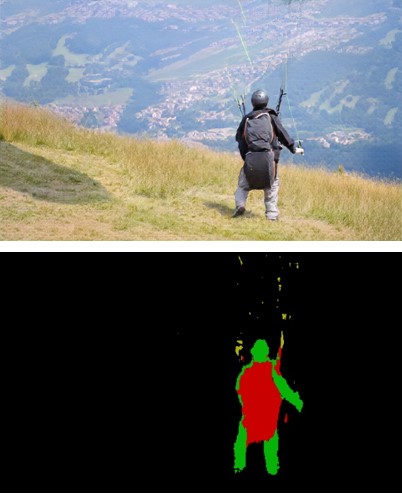} 

	\end{tabular}
	\vspace{-0.3ex}
	\caption{Trade-off between the image and mask quality when varying the number of $\mathcal{D}_{\text{low-level}}$ discriminator blocks. Increased number improves image quality, but harms the ability of masks to capture fine-grained object details due to stronger downsampling during training.		
	}
	\label{fig:granularity}
	\vspace{-1.2ex}
\end{figure*}

The quantitative comparison of the image synthesis to single-image GAN models on DAVIS-17 is presented in Table~\ref{table:ablations}. Compared to these models, OSMIS not only offers an additional ability to generate segmentation masks, but also achieves higher image quality and diversity. As seen in Table~\ref{table:ablations}, despite a potential trade-off between SIFID and LPIPS, our model outperforms previously published baselines in both metrics by a notable margin.
Further, Table~\ref{table:ablation_imagemask} demonstrates that prior image-mask methods, DatasetGAN and SemanticGAN, suffer from instabilities and fail to achieve diverse synthesis, scoring very low in LPIPS. 

\myparagraph{Ablations.}
In Table~\ref{table:comparison} we compare the proposed masked content attention module (MCA) with three alternative discriminator mechanisms to provide supervision for the generator's mask synthesis branch. The simplest baseline is to concatenate the input masks to images, requiring the discriminator to judge their realism jointly. Another method is to use projection \cite{miyato2018cgans}, by taking the inner product between the last linear layer output of $D_{\text{low-level}}$ and the pixel-wise linear projection of the input mask. Finally, we compare to the approach of SemanticGAN \cite{li2021semantic}, adding a separate discriminator network $\mathcal{D}_{m}$ which takes both segmentation masks and images, and propagate its gradients only to the generator's mask synthesis branch. While training these baselines, we preserve all the OSMIS hyperparameters, but remove the MCA and use the original $\mathcal{D}_{content}$ as in \cite{sushko2021one}.
As seen from mIoU in Table~\ref{table:comparison}, MCA enables the generation of segmentation masks with the best alignment to the generated image content, as measured by an external segmentation network. Notably, while all the alternative methods negatively affect diversity, MCA improves it (0.387 vs 0.368 LPIPS), highlighting its regularization effect which prevents the discriminator memorization of training data.

\begin{table}[t]
	
\centering
\begin{tabular}{c|c|c|c}
	
	\multirow{1}{*}{$N_{\text{low-level}}$} & \multirow{1}{*}{SIFID$\downarrow$}  & \multirow{1}{*}{LPIPS$\uparrow$} & \multirow{1}{*}{mIoU} 
	\tabularnewline
	
	\hline 	
	
	1 & \color{black}0.262\color{black} & 0.395 & \color{black}82.4\color{black}  
	\tabularnewline 
	2 & \color{black}0.165\color{black} & \textbf{0.404} & \textbf{87.1} 
	\tabularnewline
	3 & \color{black}0.102\color{black} & 0.394 & 86.9 
	\tabularnewline
	4 & 0.073 & 0.387 & 86.6 
	\tabularnewline
	5 & \textbf{0.070} & \color{black}0.321\color{black} & \color{black}83.9\color{black} 
	\tabularnewline
\end{tabular}
\vspace{1.0ex}
\caption{Ablation on the number of $\mathcal{D}_{\text{low-level}}$ discriminator blocks on DAVIS-17. 
	Bold denotes the best performance.}
\label{table:granularity} %
\vspace{-1.6ex}
\end{table}

While enabling on average higher image diversity and mask quality, we found that MCA can struggle if the training sample contains annotations of fine-grained object details, due to downsampling of input masks. This is illustrated in Fig.~\ref{fig:granularity} and Table~\ref{table:granularity}, for which we train OSMIS with different numbers of low-level discriminator blocks $N_{\text{low-level}}$, corresponding to different degrees of mask downsampling. We observe a trade-off between the quality of images and masks: decreasing $N_{\text{low-level}}$ improves the image diversity and pixel-level mask fidelity, but harms image quality. We selected $N_{\text{low-level}}=4$ as a compromise between the metrics in Table~\ref{table:granularity}, even though this configuration sometimes fails to annotate small object details (as in Fig.~\ref{fig:granularity}). Note that despite this limitation, MCA still outperforms alternative methods that do not use downsampling on DAVIS-17 (see Table~\ref{table:comparison}), and leads to image-mask pairs that are more useful as data augmentation, as discussed next.



\subsection{Application to one-shot segmentation tasks}
\label{sec:exp-applications}



After training, OSMIS can augment the provided image-mask pair with novel diverse samples. As such diversity (edited backgrounds, objects changing relative locations) is difficult to achieve by means of standard data augmentation, we foresee a potential usage of our model as a source of labelled data augmentation. Thus, in what follows, we test the efficacy of OSMIS generations when applied at test phase of two one-shot segmentation applications.


\begin{table}[t]
\centering
	
\begin{tabular}{@{\hskip 0.03in}c@{\hskip 0.03in}|@{\hskip 0.03in}c@{\hskip 0.03in}c@{\hskip 0.03in}|@{\hskip 0.04in}c@{\hskip 0.04in}@{\hskip 0.04in}c@{\hskip 0.04in}}
		\multirow{2}{*}{Network} & \multicolumn{2}{@{\hskip 0.01in}c@{\hskip 0.03in}|@{\hskip 0.04in}}{\scriptsize Augmentation:}  & \multirow{2}{*}{DAVIS-16}  & \multirow{2}{*}{DAVIS-17}
		\tabularnewline

		
		& \scriptsize Standard & \scriptsize \textbf{Ours} &  &  
		\tabularnewline	
		\hline 

		
		\multirow{4}{*}{OSVOS \cite{Cae+17}}  & \xmark & \xmark & \hspace{-2.00ex}76.9  & \hspace{-2.00ex}51.3 
		\tabularnewline
		& \cmark & \xmark  & \hspace{2.79ex}78.5~~\scriptsize{(80.2)} & \hspace{2.79ex}52.9~~\scriptsize{(52.8)}
		\tabularnewline 
		& \xmark & \cmark  & \hspace{-2.00ex}78.2  &  \hspace{-2.00ex}52.6
		\tabularnewline 
		& \cmark & \cmark  & \hspace{-2.00ex}\textbf{79.8} & \hspace{-2.00ex}\textbf{54.2}
		\tabularnewline 
		\hline 	
		
		\multirow{4}{*}{STM \cite{oh2019video}}  & \xmark & \xmark & \hspace{2.79ex}89.7~~\scriptsize{(89.4)} & \hspace{2.79ex}72.4~~\scriptsize{(72.2)}
		\tabularnewline
		& \cmark & \xmark & \hspace{-2.00ex}89.9 & \hspace{-2.00ex}72.4
		\tabularnewline 
		& \xmark & \cmark  & \hspace{-2.00ex}90.1 &  \hspace{-2.00ex}72.6
		\tabularnewline 
		& \cmark & \cmark & \hspace{-2.00ex}\textbf{90.2} & \hspace{-2.00ex}\textbf{72.7}
		
		
\end{tabular}
\vspace{1.0ex}
\caption{Effect of data augmentation on the mean of mIoU and contour accuracy ($\mathcal{J}$\&$\mathcal{F}$) of one-shot video object segmentation. Bold denotes the best performance. Round brackets show the results reported in \cite{Cae+17, oh2019video}. Reproduced and reported numbers for OSVOS differ as its official code lacks some model components.}
\label{table:downstream_video} %
\vspace{-3.2ex}
\end{table}

\myparagraph{One-shot video object segmentation.} We apply our model to the semi-supervised one-shot video segmentation benchmark DAVIS \cite{pont20172017}. At test phase, this task provides a video and the segmentation mask of objects only in the first frame, while a model is required to segment all the remaining video frames. We select two popular models from the literature: OSVOS \cite{Cae+17}, which fine-tunes the network weights on the first video frame and segments other frames independently, and STM \cite{oh2019video}, which propagates the segmentation prediction sequentially using a space-time memory module. We conduct experiments on two DAVIS splits: \textit{DAVIS-16}, having 20 videos with a single annotated object; and its extension \textit{DAVIS-17}, having 30 videos with multi-instance annotations. To evaluate the video segmentation, we compute the average of the mean mIoU region similarity ($\mathcal{J}$) and the mean contour accuracy ($\mathcal{F}$) across all videos, which is a popular metric for this task \cite{pont20172017}.

\myparagraph{One-shot semantic image segmentation.} 
The second setup is the one-shot image segmentation benchmark COCO-20$^i$ \cite{Lin2014MicrosoftCC}. 
In this task, a segmentation model is first trained on a large dataset. At test phase, the model is given a single image-mask pair (support set) with an object of a previously unseen test class, and is then required to segment another sample (query image) containing instances of the same class.  We conduct experiments with the state-of-the-art RePRI network \cite{boudiaf2021few}. COCO-20$^i$ contains 80 classes, which are divided into 4 folds, with 60 base and 20 test classes in each fold. To test OSMIS, we randomly selected 5 support samples for each test class, resulting in 100 image-mask pairs in each of the folds, and trained OSMIS on all of them separately. The performance of this task is evaluated separately for each fold, using the average mIoU across many different support-query examples.

\begin{table}[t]
\centering
\vspace{-0.5ex}
\begin{tabular}{@{\hskip 0.03in}c@{\hskip 0.03in}|@{\hskip 0.03in}c@{\hskip 0.03in}c@{\hskip 0.03in}|@{\hskip 0.04in}c@{\hskip 0.08in}c@{\hskip 0.08in}c@{\hskip 0.08in}c@{\hskip 0.04in}}
		\multirow{2}{*}{Network} & \multicolumn{2}{@{\hskip 0.02in}c@{\hskip 0.05in}|@{\hskip 0.04in}}{\scriptsize Augmentation:}  & \multirow{2}{*}{COCO$^0$} & \multirow{2}{*}{COCO$^1$} & \multirow{2}{*}{COCO$^2$} & \multirow{2}{*}{COCO$^3$}
		\tabularnewline
		
		 & \scriptsize Standard & \scriptsize \textbf{Ours} & & & & 
		\tabularnewline	
		\hline 	

		
		\multirow{5}{*}{RePRI \cite{boudiaf2021few}}  & \multirow{2}{*}{\xmark} & \multirow{2}{*}{\xmark} & 31.2 & 38.3 &  32.9  & 33.2 \vspace{-0.12cm}
		\tabularnewline
		&  &   & \scriptsize{} (31.2)~ & \scriptsize{} (38.1)~  & \scriptsize{} (33.3)~ & \scriptsize{} (33.0)~ 
		\tabularnewline \rule{0pt}{10pt}
		& \cmark & \xmark  & 31.8 & 38.5  & 33.4 & 33.8 
		\tabularnewline 
		& \xmark & \cmark  & 32.4 & 38.7  & 33.7 & 34.3 
		\tabularnewline 
		& \cmark & \cmark  & \textbf{32.8} & \textbf{39.0} & \textbf{34.1} & \textbf{34.6}
		\tabularnewline 	
		
\end{tabular}
\vspace{-0.0ex}
\caption{Effect of synthesized data augmentation on mIoU of one-shot image segmentation. In each data split, support examples were sampled from a subset of 100 image-mask pairs, for which our model was trained. Bold denotes the best performance. The round brackets contain the numbers reported in \cite{boudiaf2021few}.}
\label{table:downstream_image} %
\vspace{-2.0ex}
\end{table}

\myparagraph{Experimental setup.} For both applications, we train OSMIS on the single given image-mask pair (the first video frame or support sample). We try to closely fit the resolution of each image from COCO, and set a fixed resolution of 640x384 for images from the DAVIS benchmark. After training, we generate a pool of synthetic image-mask pairs consisting of $n=100$ samples. As OSMIS can occasionally fail and synthesize noisy examples, we compute the SIFID metric \cite{Shaham2019SinGANLA} for each generated image as a measure of its quality. Ranking the images by the average of SIFID ranks at different InceptionV3 layers, we exclude bad-quality samples by filtering out $15\%$ lowest-ranked images. Finally, we add the remaining synthetic samples to the original image-mask pair as data augmentation. See more setup details in Sec.~\ref{sec:supp_applications_setup} of the supplementary material.

Among the used segmentation models, only OSVOS~\cite{Cae+17} applies data augmentation at test phase (random combinations of image-mask flipping, zooming, and rotation). Thus, in experiments we compare our synthetic data augmentation to this pipeline (referred to as \textit{standard} augmentation).

\myparagraph{Results.} 
The performance of segmentation networks using different data augmentation is shown in Tables~\ref{table:downstream_video} and~\ref{table:downstream_image}. To account for the variance between runs, all the results are averaged across 5 runs with different seeds for augmentation. We generally managed to reproduce the official reported numbers closely, with the exception of OSVOS, for which the official codebase\footnote{https://github.com/kmaninis/OSVOS-PyTorch} does not implement the model in full configuration.
As seen in Tables~\ref{table:downstream_video} and~\ref{table:downstream_image}, the synthetic data augmentation produced by OSMIS yields a notable increase in segmentation performance, on average improving the metrics of OSVOS and STM by 1.3 and 0.3 $\mathcal{J}$\&$\mathcal{F}$ points, and RePRI by 0.9 mIoU points compared to the models using no data augmentation. 
Despite a possible mismatch between OSMIS training resolution and target image size (e.g., 640x384 vs 854x480 for DAVIS) and the need for image resizing, our synthetic data augmentation consistently outperforms standard data augmentation for STM and RePRI, and is almost on par for OSVOS, which was originally tuned for training with standard data augmentation. 
These results validate the ability of OSMIS to generate structurally diverse data augmentation of sufficient quality in the one-shot regime.
Finally, we note that the effect of OSMIS generations is complementary to standard data augmentation, as the best results for all models are observed when the two pipelines are used in combination.


Table~\ref{table:comp_augm} demonstrates the efficiency of synthetic data augmentation obtained with different GAN models. The previous image-mask models DatasetGAN and SemanticGAN both show poor applicability in the scenario of one-shot applications due to poor synthesis performance. Further, among the comparison methods for mask synthesis supervision, the strongest increase in performance is achieved with our proposed MCA module. This indicates that the high synthesis diversity and precise image-mask alignment (see Table~\ref{table:comparison}) are the keys to achieve useful data augmentation.

\begin{table}[t]
	\centering
	\vspace{-0.5ex}
	\begin{tabular}{@{\hskip 0.03in}l|@{\hskip 0.06in}c@{\hskip 0.12in}|@{\hskip 0.04in}c@{\hskip 0.03in}}
		\multirow{2}{*}{Synthesis method} & \scriptsize OSVOS, DAVIS-16 & \scriptsize RePRI, COCO$^0$
		\tabularnewline
		
		&  $\mathcal{J}$\&$\mathcal{F}$ & mIoU
		\tabularnewline	
		\hline 	
		
		\color{darkgray} Reference w/o synth. augm. &  \color{darkgray}    78.5 & \color{darkgray}  31.8
		\tabularnewline 
		\hline 
		SemanticGAN \cite{li2021semantic} & 73.1 & 29.4
		\tabularnewline 		
		DatasetGAN \cite{zhang2021datasetgan} & 77.8 & 	30.9
		\tabularnewline	
		\hline 
		
		Projection \cite{miyato2018cgans} & 78.4 & 30.9
		\tabularnewline
		Input concat. & 79.3 & 31.9
		\tabularnewline
		SemanticGAN  $D_m$ \cite{li2021semantic} & 79.5 & 32.3
		\tabularnewline
		MCA (ours) & \textbf{79.8} & \textbf{32.8}
	\end{tabular}
	\vspace{-0.0ex}
	\caption{Impact on the performance of synthesized data produced with different models and mask supervision methods. The reference performance is obtained using standard data augmentation. Bold denotes the best performance.}
	\label{table:comp_augm} %
	\vspace{-2.0ex}
\end{table}

\section{Conclusion}
\label{sec:conclusion}

We presented OSMIS, an unconditional GAN model that can learn to generate new high-quality image-mask pairs from a single training pair, not relying on any pre-training data. In such a low-data regime, our model generates photorealistic scenes that structurally differ from the original samples, while the produced masks are precisely aligned to the generated image content. Although the synthesis of OSMIS is inherently constrained by the appearance of objects in the original sample, it can serve as a source of useful data augmentation for one-shot segmentation applications, providing complementary gains to standard image augmentation. Thus, we find using one-shot image-mask synthesis in practical applications promising for future research.

\begin{flushleft}
	\small{\textbf{Aknowledgement.} Juergen Gall was supported by the Deutsche Forschungsgemeinschaft (DFG, German Research Foundation) under Germany's Excellence Strategy -- EXC 2070 -390732324 and the ERC Consolidator Grant FORHUE (101044724).}
\end{flushleft}

{\small
	\bibliographystyle{ieee_fullname}
	\bibliography{references}
}

\newpage
~

\begin{table*}[h!]
	\begin{center}
		\textbf{\Large{One-Shot Synthesis of Images and Segmentation Masks \vspace{0.5ex} \\ \textit{Supplementary material}}}
	\end{center}
\end{table*}

\newpage


\renewcommand{\thesection}{\Alph{section}}
\renewcommand{\thetable}{\Alph{table}}
\renewcommand{\thefigure}{\Alph{figure}}

\setcounter{section}{0}   
\setcounter{figure}{0}   
\setcounter{table}{0}

\section{Qualitative comparisons to prior image-mask GAN models}

\begin{figure*}[t]
	\begin{centering}
		\setlength{\tabcolsep}{0.0em}
		\par\end{centering}
	\renewcommand{\arraystretch}{1.25}
	\begin{tabular}{@{\hskip 0.02in}c@{\hskip 0.10in}c@{\hskip 0.03in}c@{\hskip 0.10in}c@{\hskip 0.03in}c@{\hskip 0.10in}c@{\hskip 0.03in}c@{\hskip 0.03in}c@{\hskip 0.03in}c}
		
		Training pair & \multicolumn{2}{c}{SemanticGAN \cite{li2021semantic}} &  \multicolumn{2}{c}{DatasetGAN \cite{zhang2021datasetgan}} & \multicolumn{3}{c}{OSMIS}
		\tabularnewline 
		
		\includegraphics[width=0.117\linewidth, height=0.14\textheight]{figures/prior_failures/ref} & 
		\includegraphics[width=0.117\linewidth, height=0.14\textheight]{figures/prior_failures/semanticgan_1} & 
		\includegraphics[width=0.117\linewidth, height=0.14\textheight]{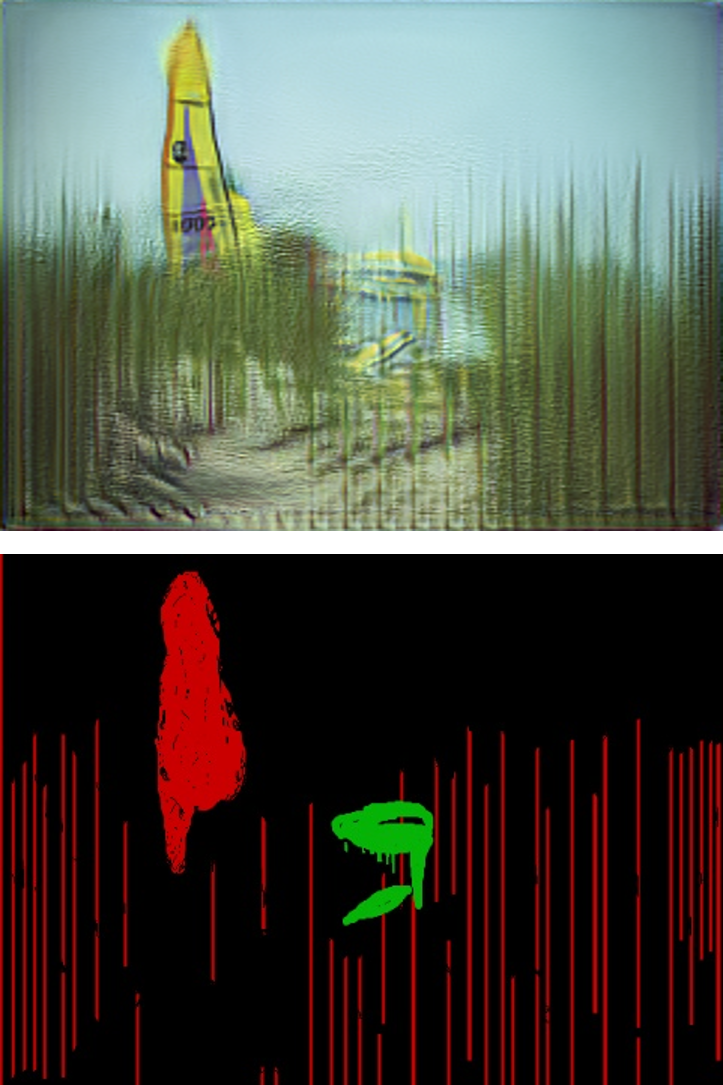} & 
		\includegraphics[width=0.117\linewidth, height=0.14\textheight]{figures/prior_failures/datasetgan_1}  &
		
		\includegraphics[width=0.117\linewidth, height=0.14\textheight]{figures/prior_failures/datasetgan_2} &
		\includegraphics[width=0.117\linewidth, height=0.14\textheight]{supplementary/figures/suppl_prior_failures/canvas/59} & 
		\includegraphics[width=0.117\linewidth, height=0.14\textheight]{supplementary/figures/suppl_prior_failures/canvas/56} & 
		\includegraphics[width=0.117\linewidth, height=0.14\textheight]{supplementary/figures/suppl_prior_failures/canvas/35}
		\tabularnewline \vspace{0.8ex}
		
		\includegraphics[width=0.117\linewidth, height=0.14\textheight]{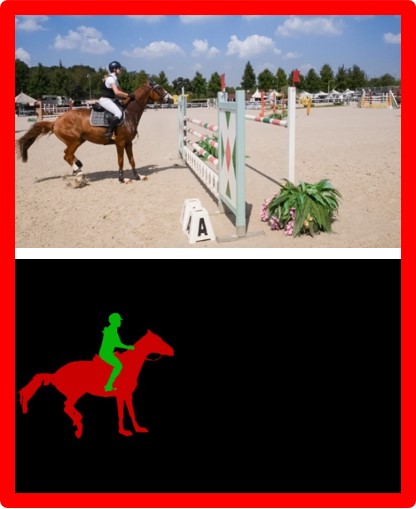} & 
		\includegraphics[width=0.117\linewidth, height=0.14\textheight]{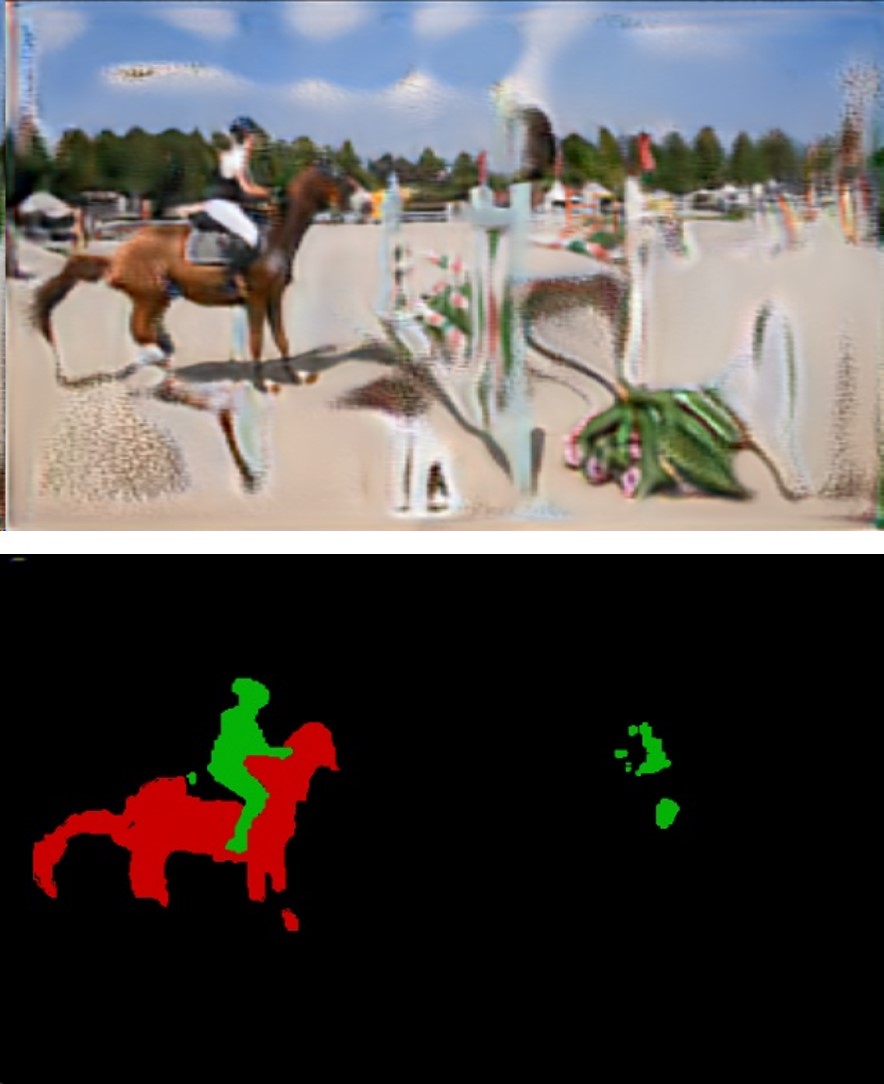} & 
		\includegraphics[width=0.117\linewidth, height=0.14\textheight]{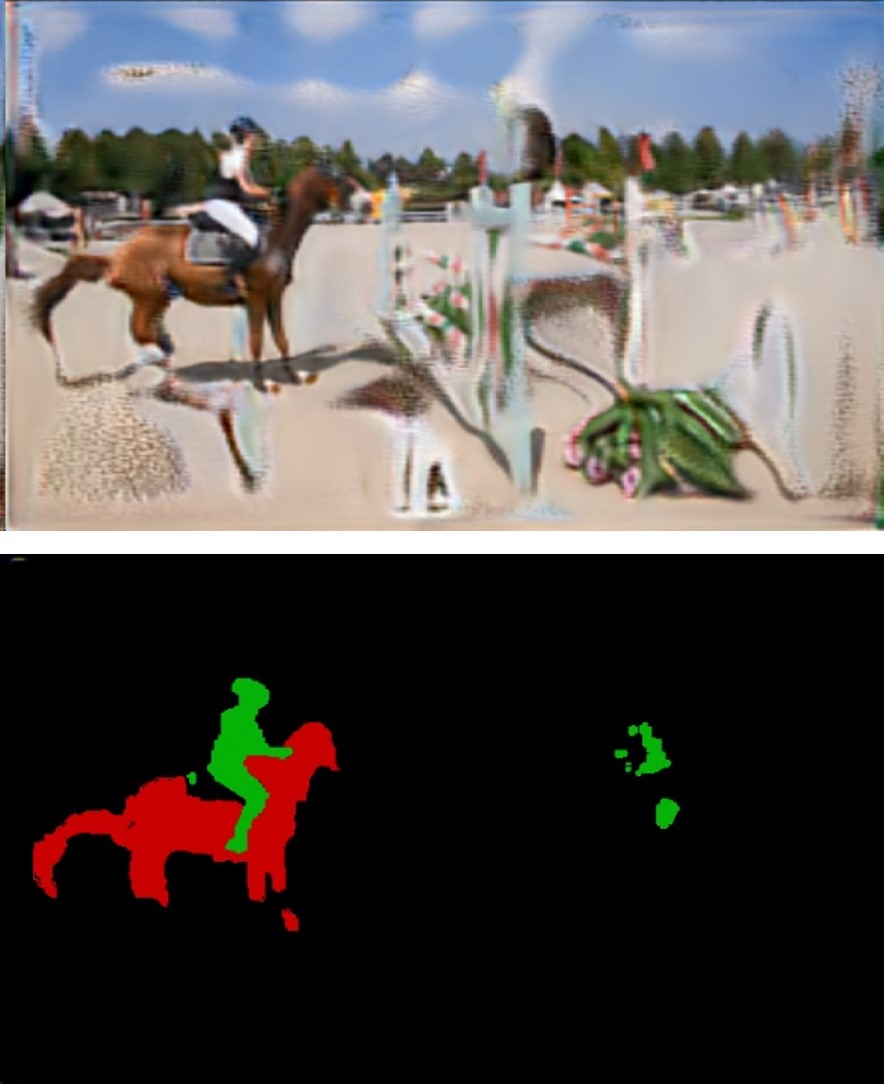} & 
		\includegraphics[width=0.117\linewidth, height=0.14\textheight]{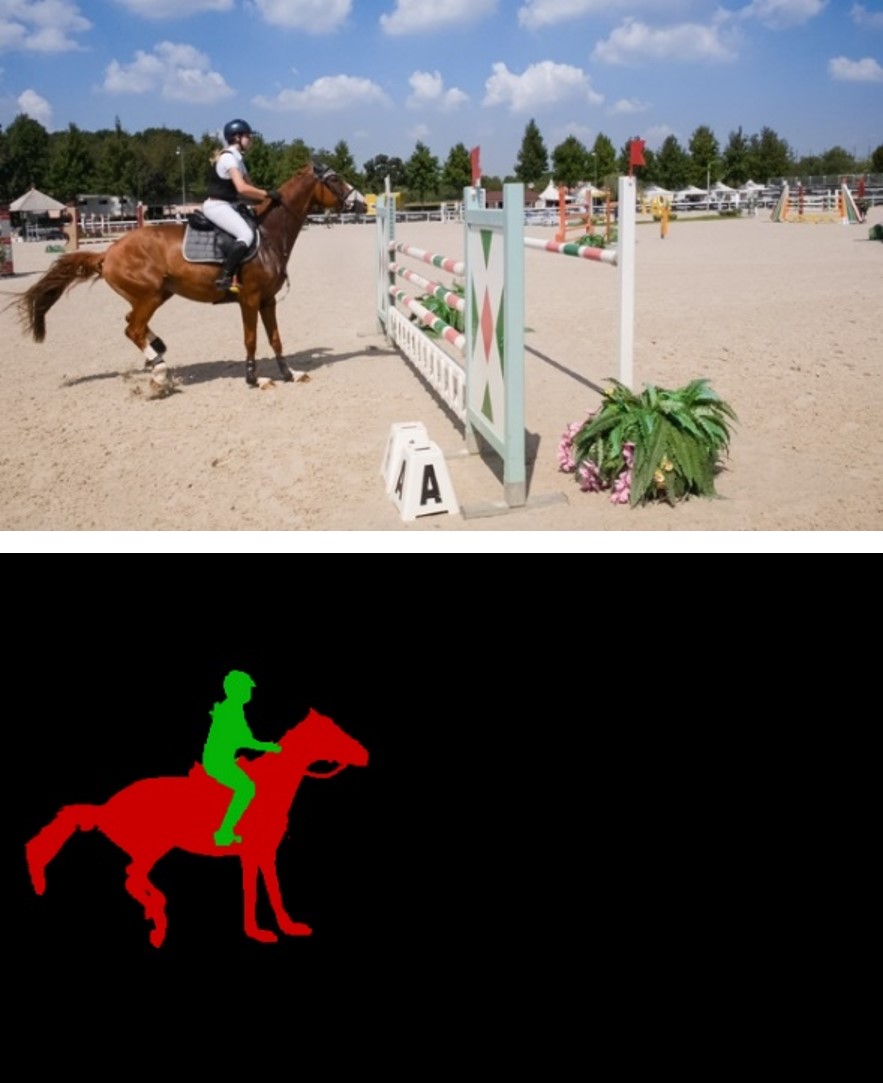}  &
		
		\includegraphics[width=0.117\linewidth, height=0.14\textheight]{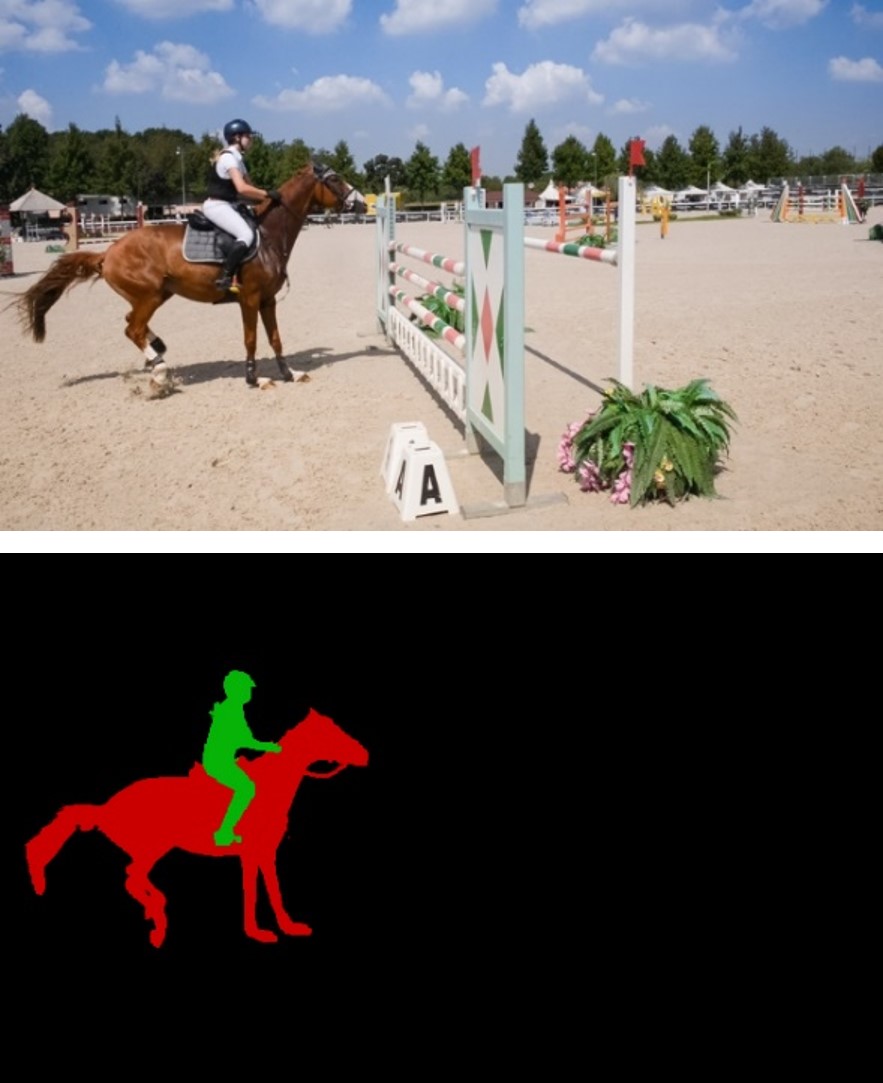} &
		\includegraphics[width=0.117\linewidth, height=0.14\textheight]{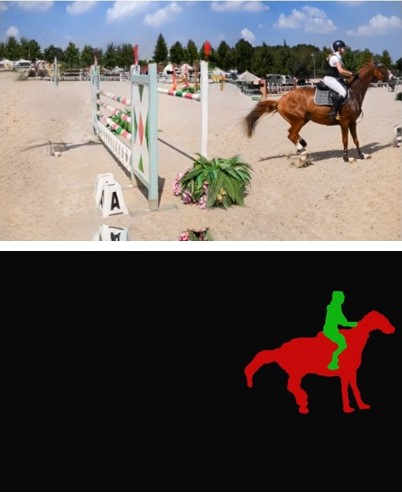} & 
		\includegraphics[width=0.117\linewidth, height=0.14\textheight]{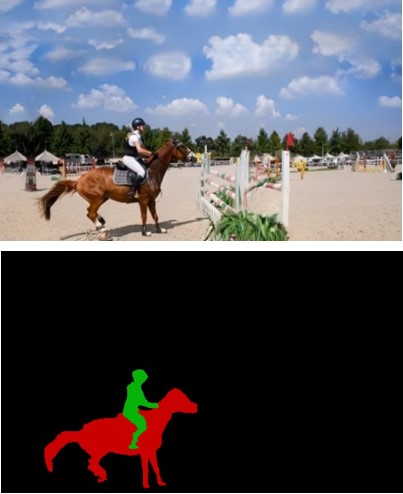} & 
		\includegraphics[width=0.117\linewidth, height=0.14\textheight]{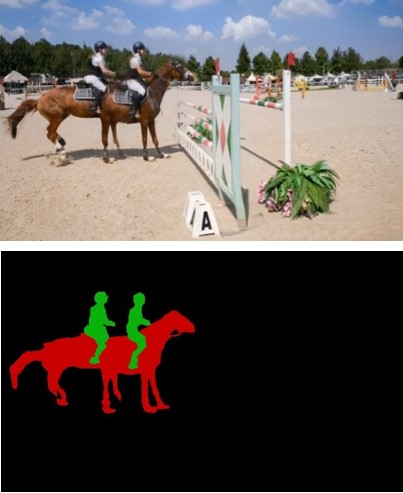}
		 
		\tabularnewline \vspace{0.8ex}
		
				\includegraphics[width=0.117\linewidth, height=0.14\textheight]{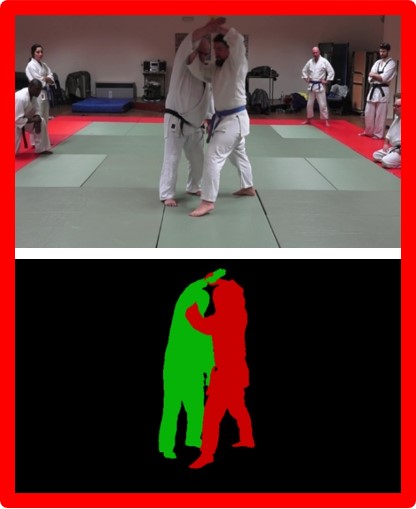} & 
		\includegraphics[width=0.117\linewidth, height=0.14\textheight]{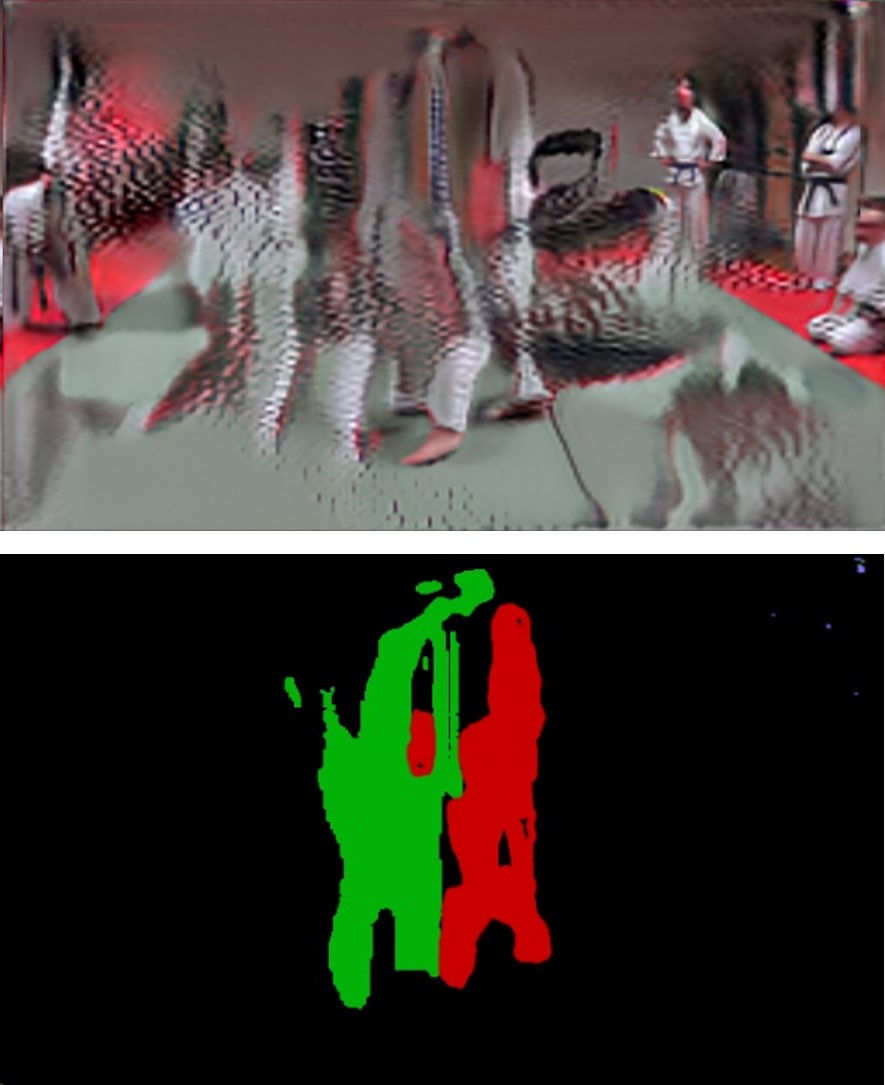} & 
		\includegraphics[width=0.117\linewidth, height=0.14\textheight]{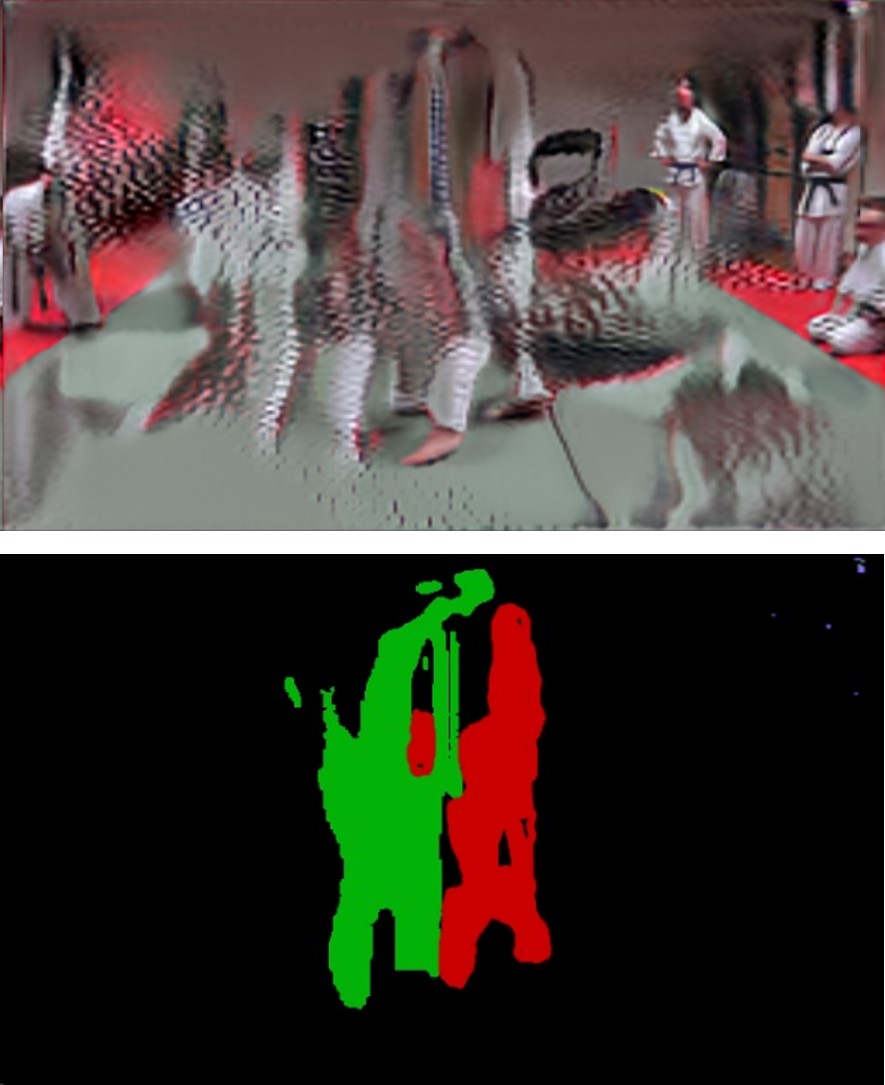} & 
		\includegraphics[width=0.117\linewidth, height=0.14\textheight]{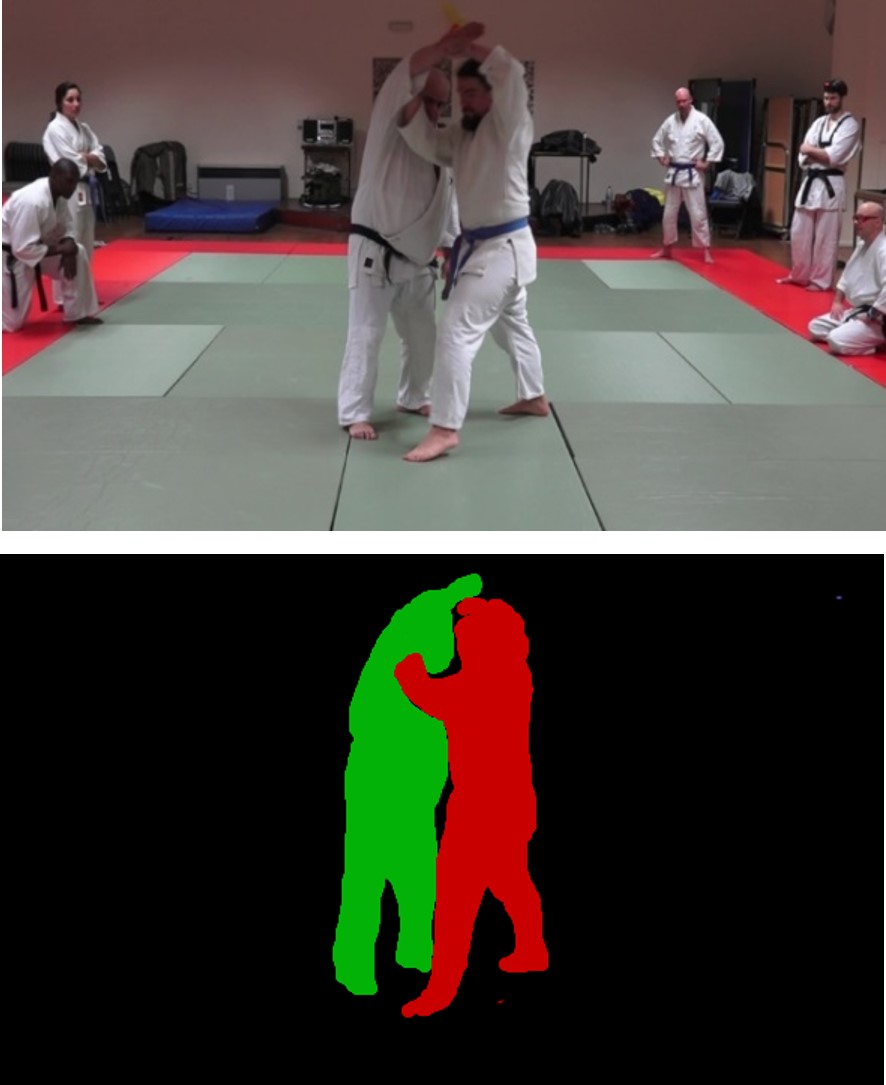}  &
		
		\includegraphics[width=0.117\linewidth, height=0.14\textheight]{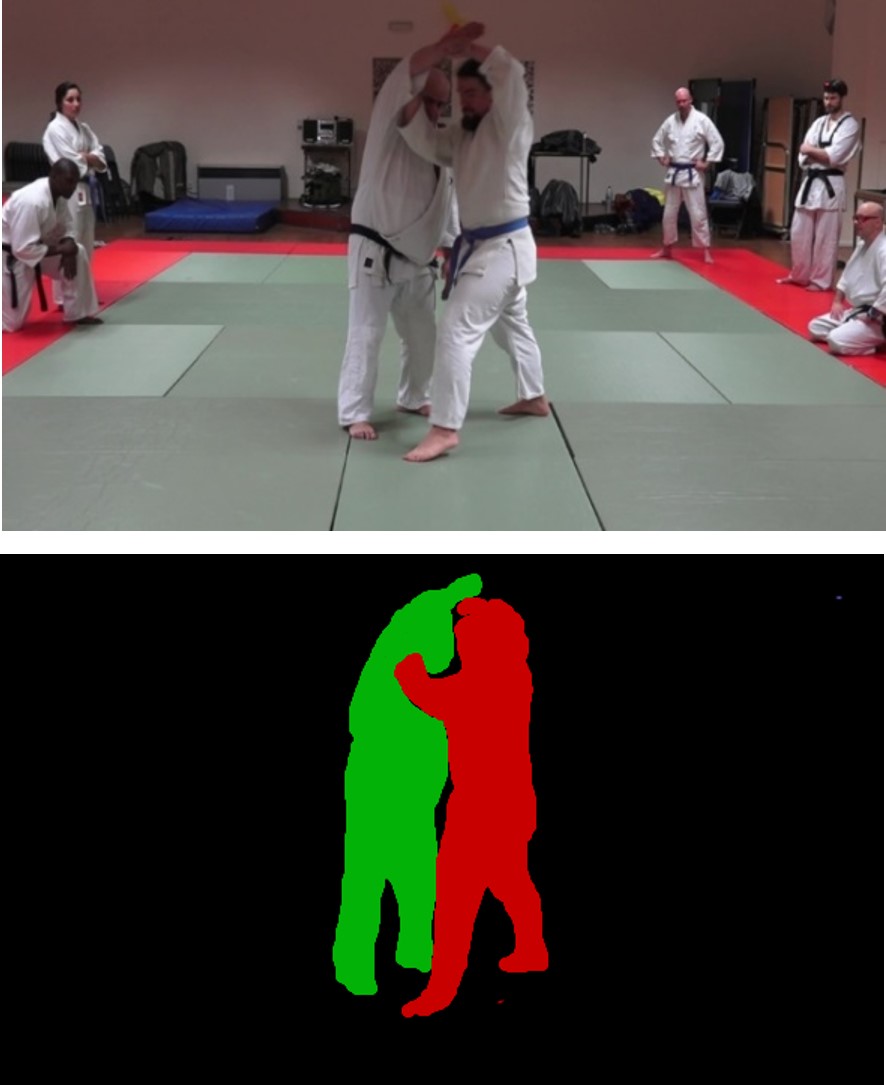} &
		\includegraphics[width=0.117\linewidth, height=0.14\textheight]{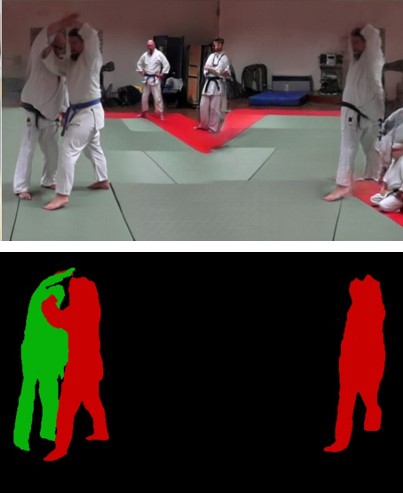} & 
		\includegraphics[width=0.117\linewidth, height=0.14\textheight]{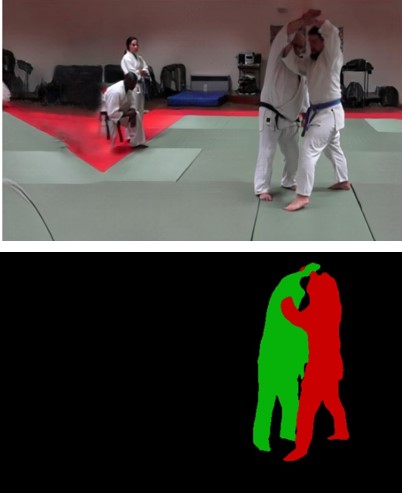} & 
		\includegraphics[width=0.117\linewidth, height=0.14\textheight]{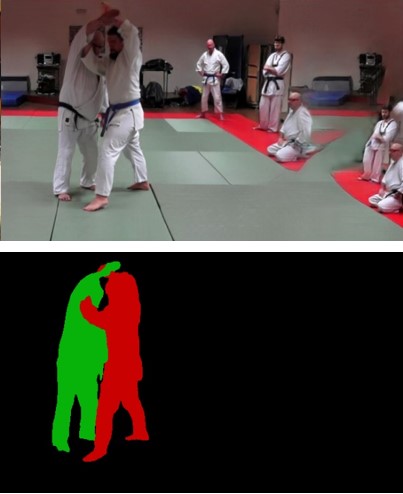}
		
		\tabularnewline \vspace{0.8ex}
		
				\includegraphics[width=0.117\linewidth, height=0.14\textheight]{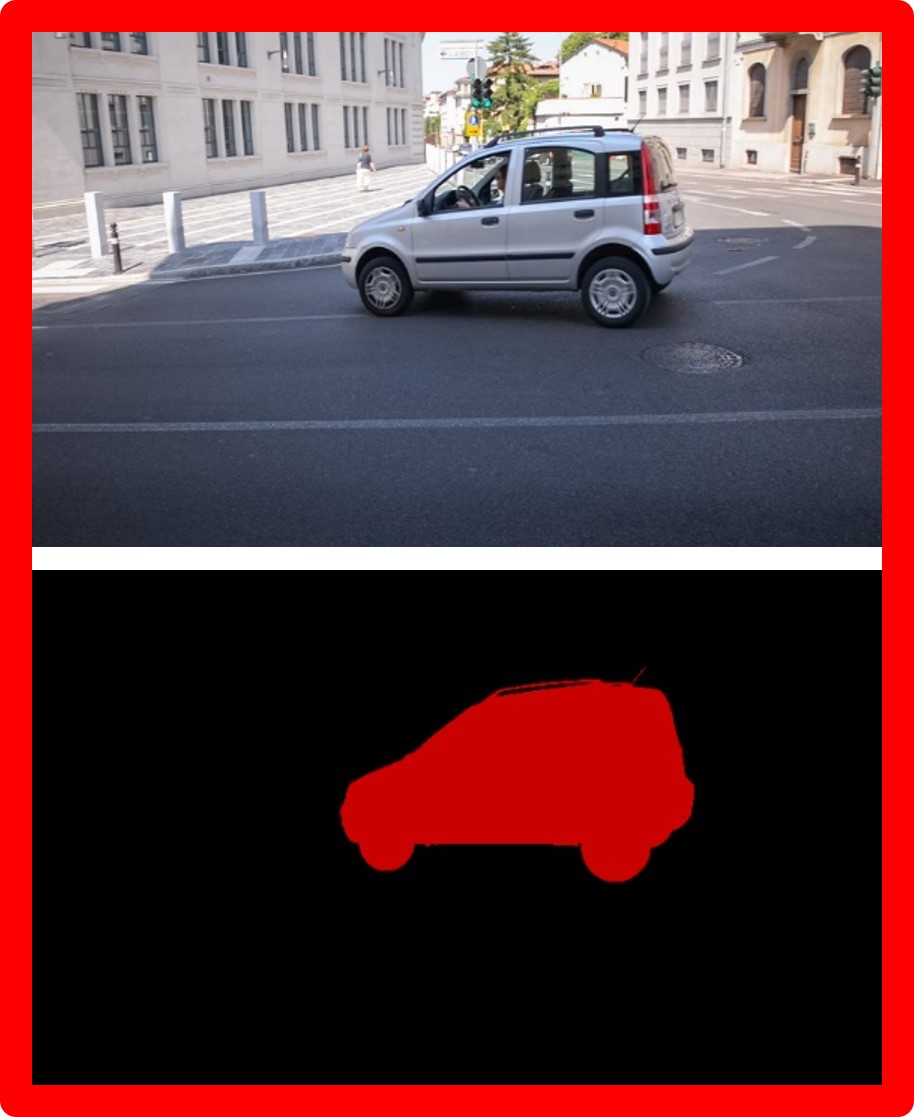} & 
		\includegraphics[width=0.117\linewidth, height=0.14\textheight]{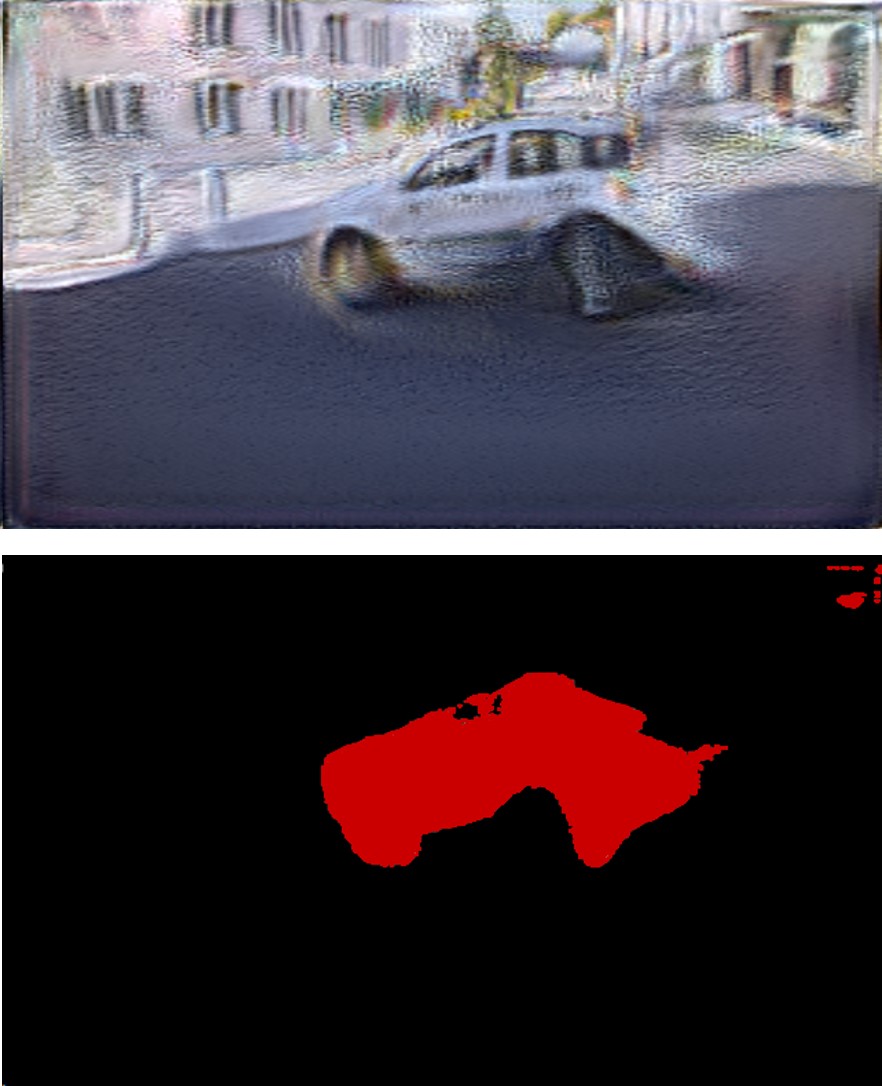} & 
		\includegraphics[width=0.117\linewidth, height=0.14\textheight]{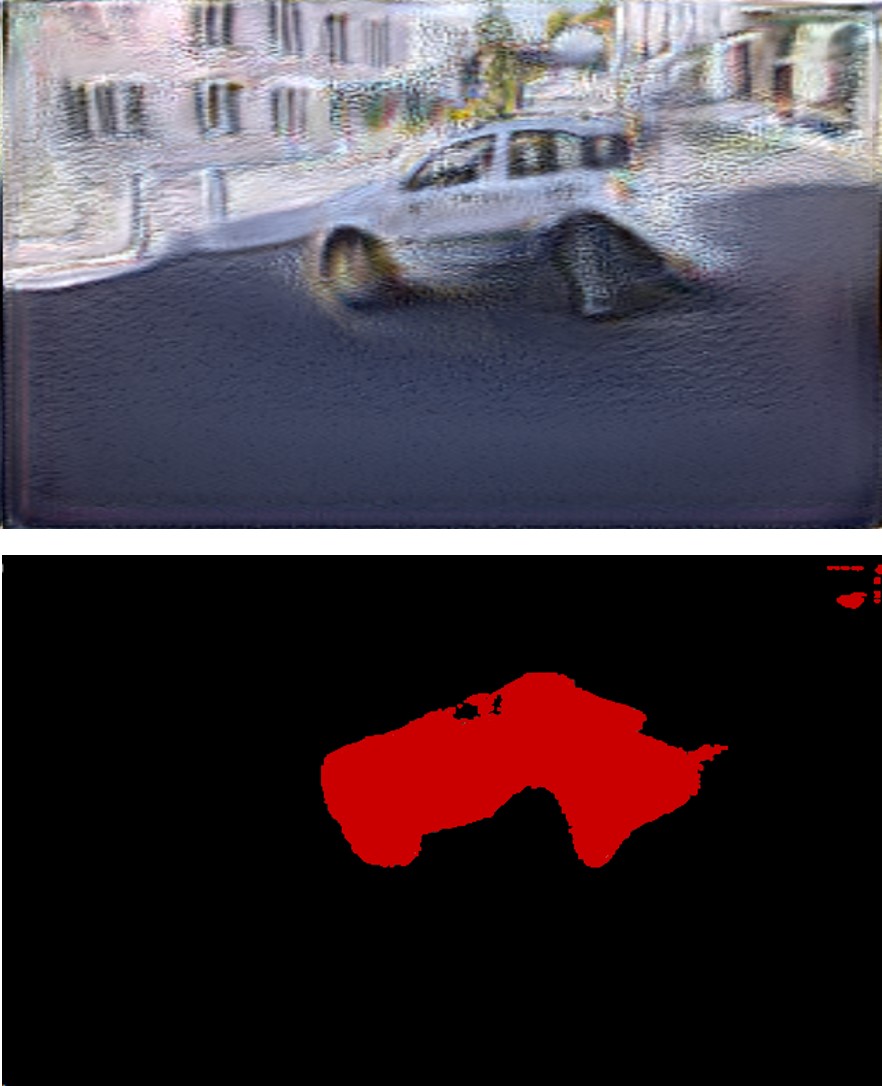} & 
		\includegraphics[width=0.117\linewidth, height=0.14\textheight]{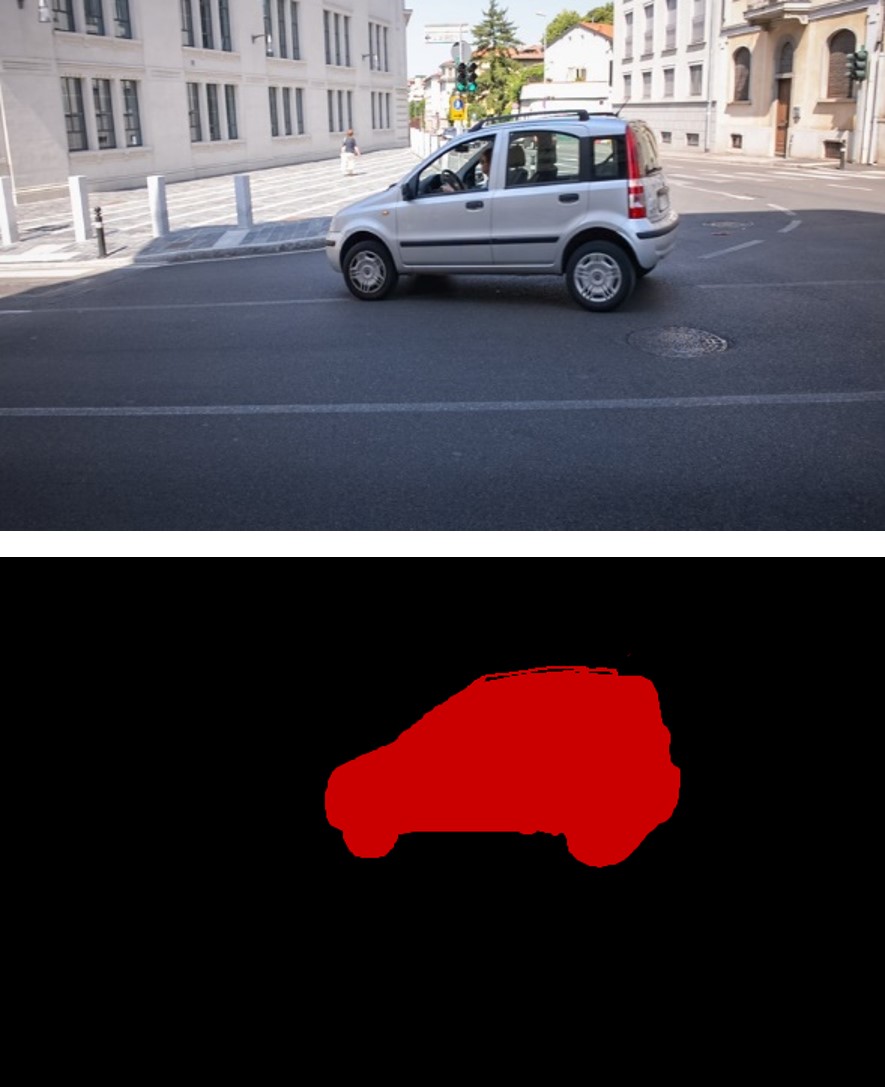}  &
		
		\includegraphics[width=0.117\linewidth, height=0.14\textheight]{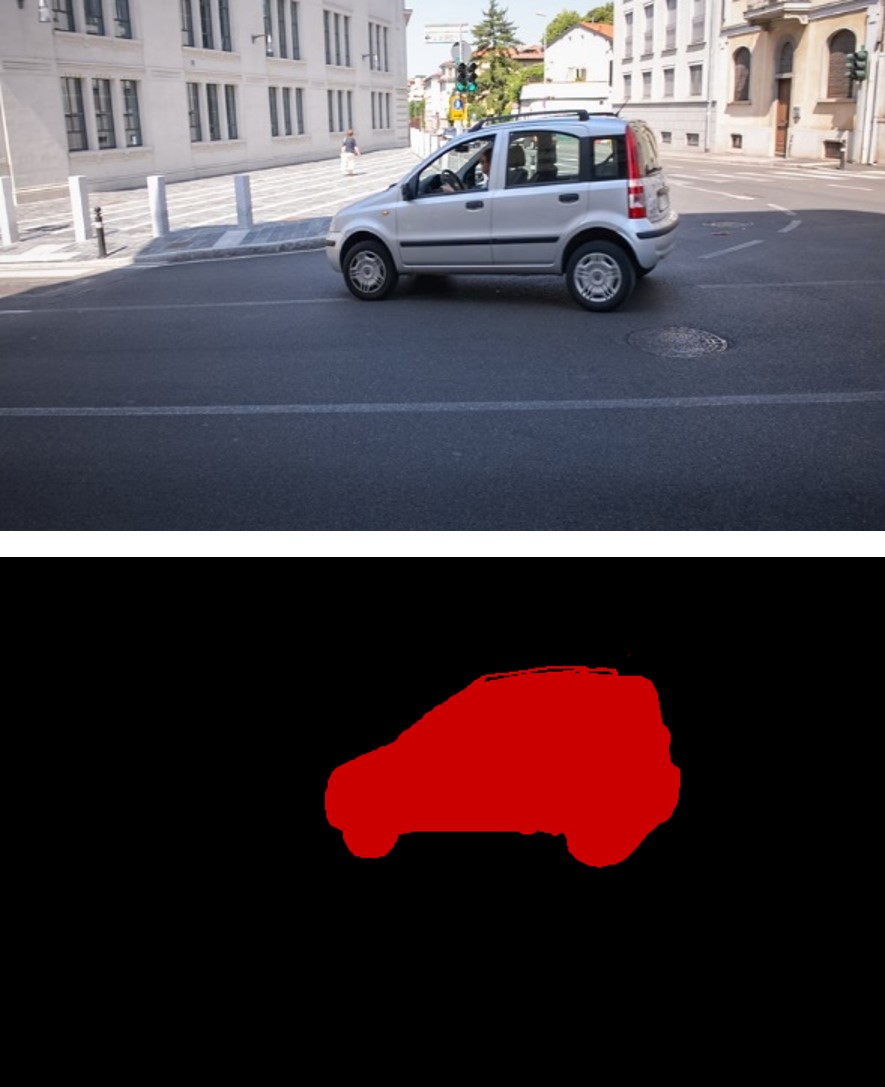} &
		\includegraphics[width=0.117\linewidth, height=0.14\textheight]{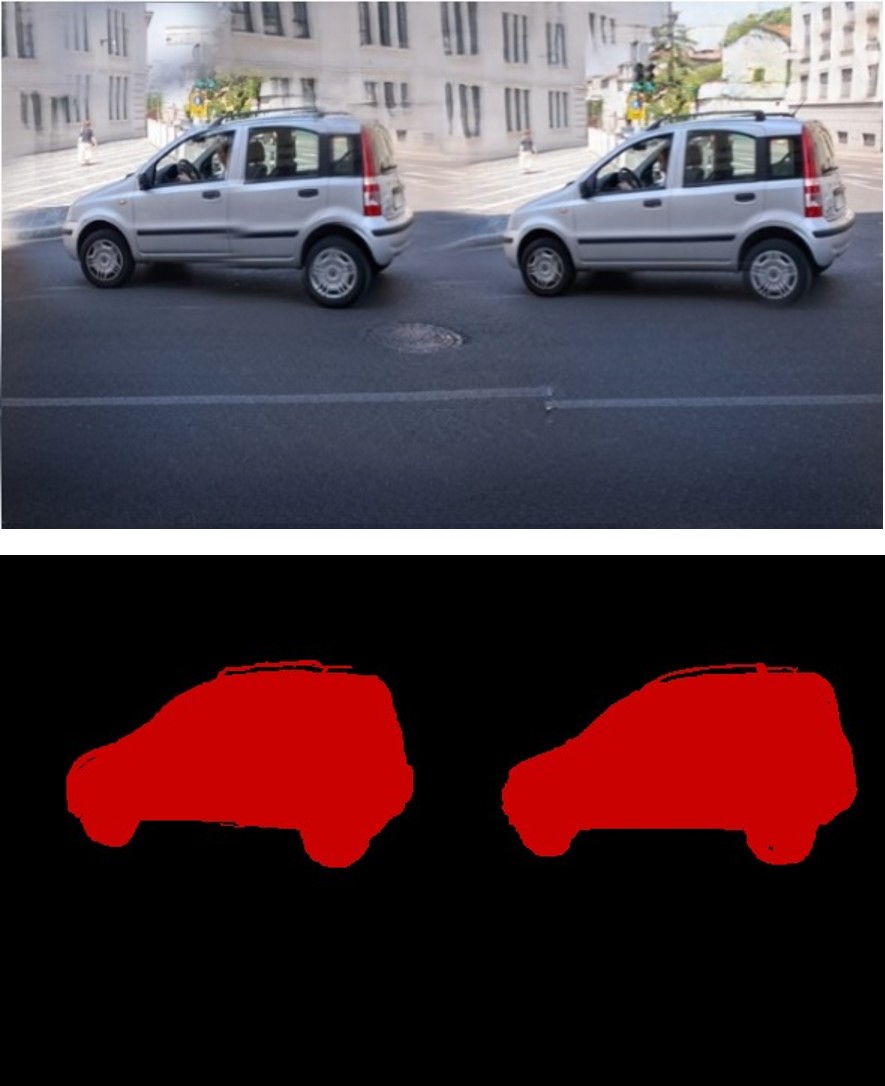} & 
		\includegraphics[width=0.117\linewidth, height=0.14\textheight]{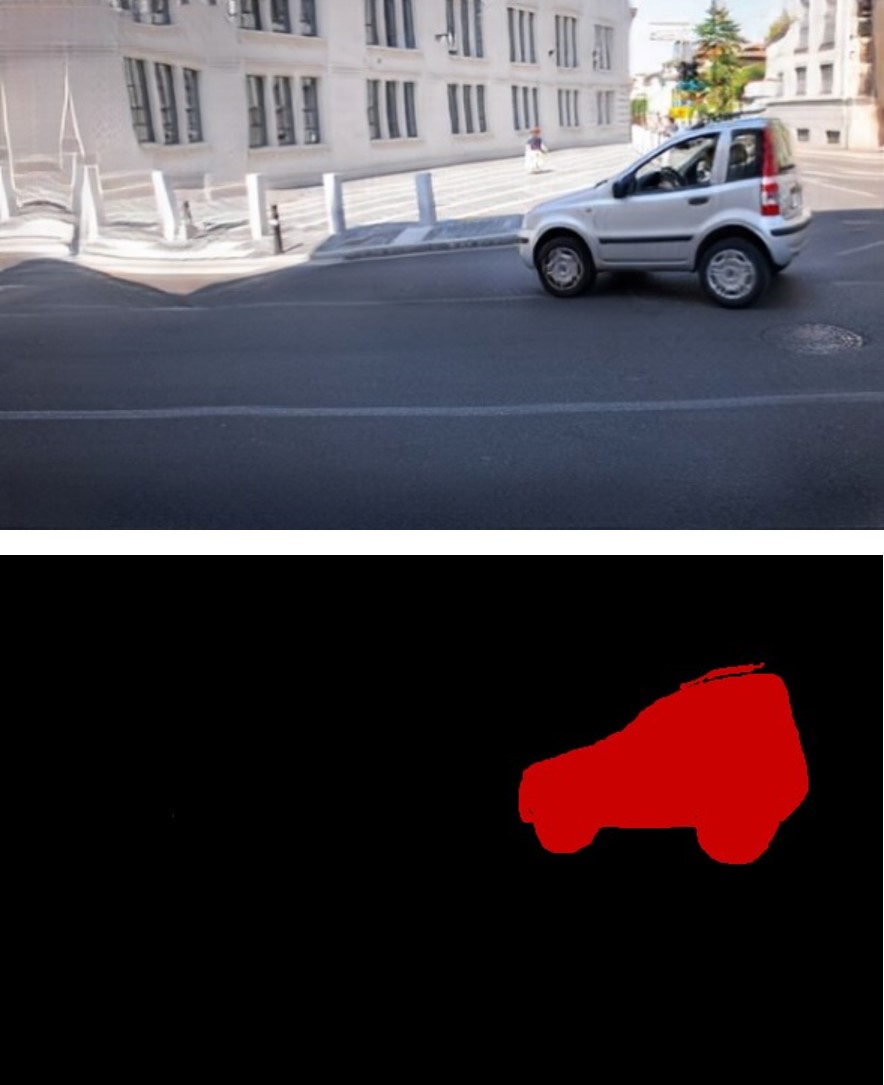} & 
		\includegraphics[width=0.117\linewidth, height=0.14\textheight]{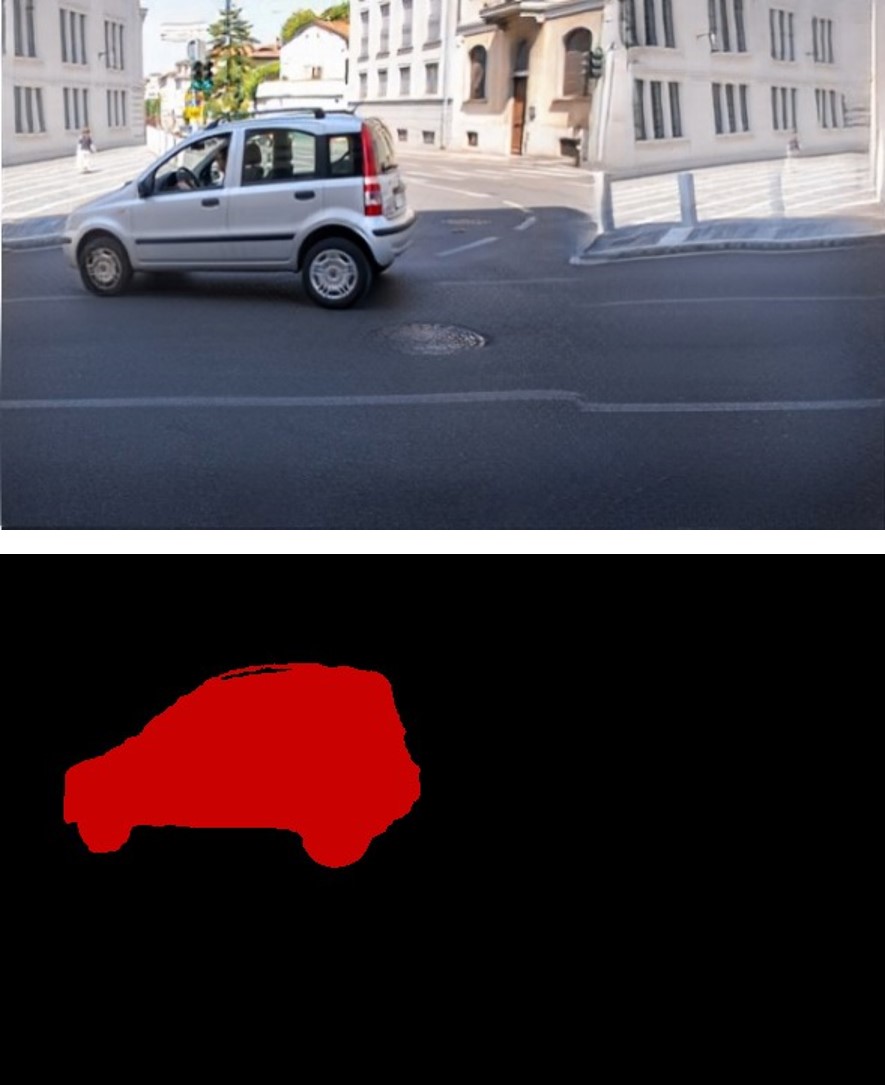}		
	\end{tabular}
	\caption{A quantitative comparison of OSMIS to previous image-mask GAN models SemanticGAN~\cite{li2021semantic} and DatasetGAN~\cite{zhang2021datasetgan}. Both the models suffer from memorization, repeating the layout of the training samples, while SemanticGAN also achieves poor visual quality of images and masks due to training instabilities. In contrast, OSMIS achieves both diversity and quality, placing foreground objects in different locations in the scene and editing the layouts of backgrounds.}
	\label{supp_fig:prior_failures}
\end{figure*}

A qualitative comparison of OSMIS to prior image-mask GAN models, SemanticGAN~\cite{li2021semantic} and DatasetGAN~\cite{zhang2021datasetgan}, is presented in Fig.~\ref{supp_fig:prior_failures}, corresponding to the quantitative comparison of these models from Table~\ref{table:ablation_imagemask}. The displayed samples were generated with a checkpoint that achieved the lowest SIFID \cite{Shaham2019SinGANLA}. Like OSMIS, SemanticGAN was trained from scratch, using a single provided image-mask pair as real data. On the other hand, the training of DatasetGAN consisted of two stages: pre-training of the StyleGAN \cite{Karras2018ASG} backbone architecture on the single provided training image, and training a label synthesis branch with manual segmentation annotations of generated images. In our one-shot setup, since StyleGAN typically collapsed to generating the same image, annotating a single generated sample was enough to train the label synthesis branch.

As seen from Figure~\ref{supp_fig:prior_failures}, both SemanticGAN and DatasetGAN suffer from memorization issues, always producing the same image that repeats the layout of the training sample. In Table~\ref{table:ablation_imagemask} this is reflected in very low LPIPS diversity scores achieved by both models. In addition, SemanticGAN shows unstable training in our one-shot regime, which results in a low visual quality of generated images and noisy annotations (note poor performance in SIFID and mIoU in Table~\ref{table:ablation_imagemask}). For DatasetGAN, we observed no such instabilities, which made the manual annotation of generated images straightforward. Despite a good visual image quality and accurate manual annotation of masks (high mIoU in Table~\ref{table:ablation_imagemask}), the low diversity of DatasetGAN prevents it from producing useful data augmentation for one-shot segmentation tasks (see Table~\ref{table:comp_augm}).

In contrast, OSMIS achieves high diversity and visual quality of generated image-masks at the same time. For example, in the examples from Fig.~\ref{supp_fig:prior_failures} our model can change the number of sails, horse riders, sumo wrestlers, or cars, at the same time editing the layout of the backgrounds, while still preserving the realism of objects. Such structural diversity of OSMIS enables its effective generation of data augmentation for one-shot segmentation tasks (see Sec~\ref{sec:exp-applications}).

\section{Additional details on the application of OSMIS to one-shot segmentation tasks}
\label{sec:supp_applications_setup}

\subsection{Details of the experimental setup}

Tables~\ref{table:downstream_video} and~\ref{table:downstream_image} show the performance of one-shot segmentation networks using different data augmentation strategies. The simplest strategy is to use no data augmentation, when the fine-tuning of networks is performed only on a single provided image-mask pair. When fine-tuning with our synthesized data augmentation, we extend the pool of the available data with $85$ filtered samples generated by OSMIS. Finally, when adding standard data augmentation to the two previous strategies, we apply random combinations of image-mask flipping, zooming, and rotation to the samples from the pool. The exact method of utilizing data augmentation depends on the segmentation network, as described next.

\myparagraph{OSVOS \cite{Cae+17}} fine-tunes weights of a pre-trained segmentation network on the image and mask of the first frame of a given video sequence. At each fine-tuning epoch, we double the batch size and randomly add generated image-mask pairs to the original data. Therefore, we keep the 50\%-50\% ratio between real and synthetic data, which we found to yield the best video segmentation performance.

\myparagraph{STM \cite{oh2019video}} scans a given video sequence frame-by-frame, starting from the first frame, for which a mask annotation is provided. This image-mask pair, as well as each K-th pair of a video frame and its segmentation prediction are added to a spatio-temporal memory bank. The memory bank is used to make the segmentation prediction of the latest video frames more accurate. To employ data augmentation, we added synthesized image-mask pairs to the STM memory bank at step 0, before processing the first video frame. To fit the memory bank into GPU memory, we had to limit the number of added samples to 10, which were sampled randomly from the synthetic pool.

\myparagraph{RePRI \cite{boudiaf2021few}} trains a small pixel-level classifier given a single support image-mask pair containing an object of a previously unseen class. We simply provide synthetic image-mask pairs as data augmentation for the original data. To fit the extended support set into GPU memory, we limited the number of added samples to 10. This way, the task of RePRI could be technically regarded as 11-shot semantic image segmentation, where all the available support data originates from a provided data sample.

\subsection{Ablation on filtering out bad-quality samples}

Filtering out noisy synthetic examples before forming a pool of synthetic samples is an important step to achieve good performance of data augmentation. For example, using generated image-mask pairs without filtering resulted in modest or negative performance gains for one-shot segmentation networks (see Table 
\ref{table:rejections}). On the contrary, a simple strategy to filter out 15\% of lowest-ranked generated images by SIFID, computed after the first pooling layer of the InceptionV3 network, helps to reduce the impact of bad-quality augmentation and, in effect, substantially improves the final segmentation performance. 

\begin{figure*}[t]
	\begin{centering}
		\setlength{\tabcolsep}{0.0em}
		\par\end{centering}
	\renewcommand{\arraystretch}{1.25}
	\begin{tabular}{c@{\hskip 0.02in}c@{\hskip 0.06in}c@{\hskip 0.04in}c@{\hskip 0.04in}c@{\hskip 0.04in}c@{\hskip 0.04in}c}

	& &  Best & Top-10\% & Top-50\% & Top-90\% & Worst
 	\tabularnewline
		
	\multirow{8}{*}{\begin{tabular}{c} \hspace{-0.4em} Training image \\ 	\includegraphics[width=0.15\linewidth, height=0.070\textheight]{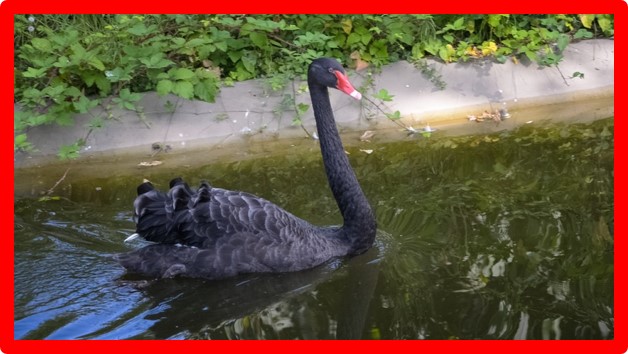} \end{tabular} }&
	
	\rotatebox{90}{~~~SIFID-1~} & 
	\includegraphics[width=0.15\linewidth, height=0.070\textheight]{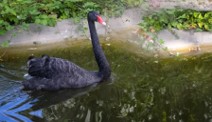} & 
	\includegraphics[width=0.15\linewidth, height=0.070\textheight]{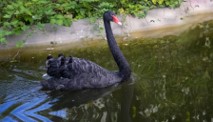} & 
	\includegraphics[width=0.15\linewidth, height=0.070\textheight]{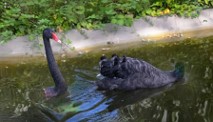} & 
	\includegraphics[width=0.15\linewidth, height=0.070\textheight]{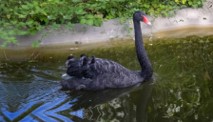}  &
	\includegraphics[width=0.15\linewidth, height=0.070\textheight]{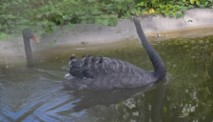}
	\tabularnewline
		
	&
	\rotatebox{90}{~~~SIFID-2~} & 
	\includegraphics[width=0.15\linewidth, height=0.070\textheight]{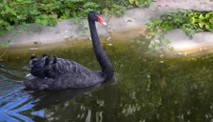} & 
	\includegraphics[width=0.15\linewidth, height=0.070\textheight]{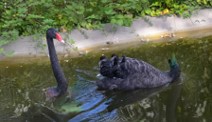} & 
	\includegraphics[width=0.15\linewidth, height=0.070\textheight]{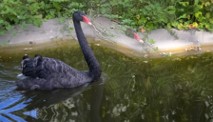} & 
	\includegraphics[width=0.15\linewidth, height=0.070\textheight]{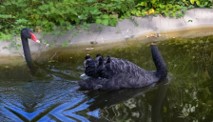}  &
	\includegraphics[width=0.15\linewidth, height=0.070\textheight]{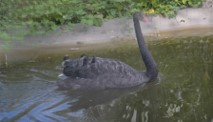}
	\tabularnewline

	&
	\rotatebox{90}{~~~SIFID-3~} & 
	\includegraphics[width=0.15\linewidth, height=0.070\textheight]{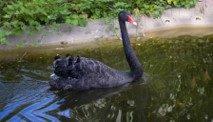} & 
	\includegraphics[width=0.15\linewidth, height=0.070\textheight]{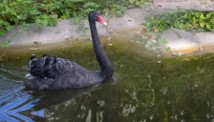} & 
	\includegraphics[width=0.15\linewidth, height=0.070\textheight]{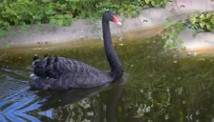} & 
	\includegraphics[width=0.15\linewidth, height=0.070\textheight]{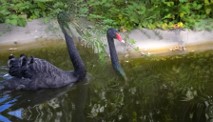}  &
	\includegraphics[width=0.15\linewidth, height=0.070\textheight]{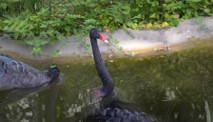}
	\tabularnewline

	&
	\rotatebox{90}{~~~SIFID-4~} & 
	\includegraphics[width=0.15\linewidth, height=0.070\textheight]{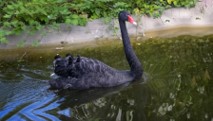} & 
	\includegraphics[width=0.15\linewidth, height=0.070\textheight]{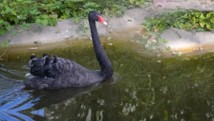} & 
	\includegraphics[width=0.15\linewidth, height=0.070\textheight]{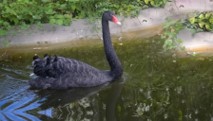} & 
	\includegraphics[width=0.15\linewidth, height=0.070\textheight]{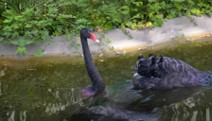}  &
	\includegraphics[width=0.15\linewidth, height=0.070\textheight]{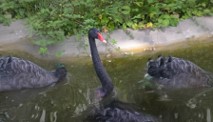}
	\tabularnewline
	\end{tabular}
	\caption{Generated images shown for different levels of SIFID, computed at various InceptionV3 layers. We observed that SIFID at the earliest InceptionV3 layers is biased towards low-level image statistics, such as colors and small textures, and is not indicative of image quality at higher scales (appearance of objects, layout of backgrounds). Thus, to filter out noisy generated examples, we use a joint ranking of images at different InceptionV3 layers.}
	\label{supp_fig:filtering}
	\vspace{-1.5ex}
\end{figure*}

However, we observed that the SIFID metric is biased towards low-level image statistics, such as color and texture distributions, and is not indicative of the quality of generated images at higher scales. We illustrate this in Fig.~\ref{supp_fig:filtering}, where we display visual examples of images at different levels of SIFID, obtained after the first pooling layer, second pooling layer, pre-classifier features, and the final features of the InceptionV3 network (denoted as SIFID-1,2,3,4). 

To account for the quality of generated images at different scales, we ranked synthesized examples by a joint ranking, taking the average of their ranks across different SIFIDs. As seen in Table~\ref{table:rejections}, filtering out noisy examples using this strategy helps to boost the performance of one-shot segmentation networks. Furthermore, we observed that it helps to significantly decrease the performance variance between different runs, which generally increased while using synthetic data augmentation in our experiments.

\begin{table}[t]
	
	\begin{tabular}{@{\hskip -0.02in}lc@{\hskip 0.03in}|@{\hskip 0.03in}c@{\hskip 0.03in}|@{\hskip 0.03in}c@{\hskip 0.03in}}
		\multirow{2}{*}{Data selection} & \multirow{2}{*}{$\eta$} & \scriptsize OSVOS, DAVIS-16 & \scriptsize RePRI, COCO$^0$
		\tabularnewline
		
		& &   $\mathcal{J}$\&$\mathcal{F}$ & mIoU
		\tabularnewline	
		\hline 	
		
		\multicolumn{2}{@{\hskip -0.02in}l@{\hskip 0.03in}|@{\hskip 0.03in}}{\color{darkgray} Reference w/o augmentations} &  \color{darkgray}    78.5 \scriptsize{(+0.0)} \scriptsize{$\pm$0.3} & \color{darkgray}  31.2 \scriptsize{(+0.0)} \scriptsize{$\pm$0.1}
		\tabularnewline 
		\arrayrulecolor{lightgray} \hline \arrayrulecolor{black}
		
		No data selection & - &     78.7 \scriptsize{(+0.2)} \scriptsize{$\pm$0.6} &  30.7 \scriptsize{(-0.5)} \scriptsize{$\pm$0.5}
		\tabularnewline 
		\arrayrulecolor{lightgray} \hline \arrayrulecolor{black}

		Only SIFID-pool$_1$ & 15\% &       79.3 \scriptsize{(+0.8)} \scriptsize{$\pm$0.5} &  32.2 \scriptsize{(+1.0)} \scriptsize{$\pm$0.4}
		\tabularnewline 
		\arrayrulecolor{lightgray} \hline \arrayrulecolor{black}

		\multirow{5}{*}{SIFID-\{1,2,3,4\} (ours)} & 5\%  &  79.3 \scriptsize{(+0.8)} \scriptsize{$\pm$0.6} &  31.9 \scriptsize{(+0.7)} \scriptsize{$\pm$0.4}
		\tabularnewline
		& 10\% &       79.6 \scriptsize{(+1.1)} \scriptsize{$\pm$0.4} &  \underline{32.6 \scriptsize{(+1.4)}} \scriptsize{$\pm$0.2}
		\tabularnewline
		& 15\% &      \textbf{79.8 \scriptsize{(+1.3)}} \scriptsize{$\pm$0.3} & \textbf{32.8 \scriptsize{(+1.6)}} \scriptsize{$\pm$0.2}
		\tabularnewline
		& 25\% &       \underline{79.7 \scriptsize{(+1.2)}} \scriptsize{$\pm$0.3} &  32.3 \scriptsize{(+1.1)} \scriptsize{$\pm$0.2}
		\tabularnewline
		& 50\% &    79.5 \scriptsize{(+1.0)} \scriptsize{$\pm$0.3} &  32.0 \scriptsize{(+0.9)} \scriptsize{$\pm$0.1}	
		\tabularnewline


	\end{tabular}
	\caption{Impact of synthetic data selection strategies on one-shot segmentation performance. Bold and underlined show the first and second best performance.}
	\label{table:rejections} %
	\vspace{-1.5ex}
\end{table}

Finally, we conduct an ablation on how many lowest-ranked images should be filtered for optimal performance. Table~\ref{table:rejections} demonstrates that the filtering rate should be neither too low nor too high: filtering out only 5\% or 10\% leaves some low quality images that are harmful for the data augmentation efficiency, while filtering too many samples (25\%, 50\%) decreases the diversity of the synthetic data pool and thus also diminishes its effectiveness.

Overall, we conclude that data filtering is a crucial step that is needed to achieve high performance gains with the help of synthetic data augmentation. Table~\ref{table:rejections} shows that our proposed data selection scheme is effective at filtering out bad generated examples, which results in higher performance of one-shot segmentation networks without notably increasing their variance between runs.

\section{Architecture of OSMIS and training details}

The architecture of the OSMIS generator and discriminator is summarized in Tables~\ref{supp_table:g_arch} and \ref{supp_table:d_arch}. We build upon the structure of One-Shot GAN \cite{sushko2021one}, which utilizes ResNet blocks for both the generator and discriminator, enables multi-scale gradients (MSG) \cite{karnewar2019msg} by employing skip connections between the latest generator layers and the low-level discriminator $D_{low-level}$, and provides control over the final image resolution by changing the input noise shape.

To achieve image-mask synthesis at a high resolution of 384x640, we set the input noise shape to 3$\times$5, use $8$ ResNet blocks in the generator, $4$ ResNet blocks for the low-level discriminator $D_{low-level}$, and $4$ blocks for the object and layout discriminators  $D_{object}$ and $D_{layout}$. Before feeding the intermediate features $F(x) = D_{low-level}(x)$ of an input image $x$ to $D_{object}$, we process it by the masked content attention module (MCA), which forms $N$ content representations, corresponding to objects or background in the image. Thus, for the object discriminator we use a batch size which is $N$ times higher than for other discriminator parts.

We train OSMIS with the ADAM optimizer \cite{adamopt}, using a batch size of 3, momenta $(\beta_1, \beta_2) = (0.5, 0.999)$, and a learning rate of 0.0002. During training, we use an exponential moving average of the generator weights with a decay of 0.9999, which is used at inference. $P_0$ from Eq.~\eqref{eq:loss_content} is set to $15000$ epochs. We extend the differentiable augmentation (DA) pipeline used in \cite{sushko2021one} by using the whole set of transformations as proposed in \cite{Karras2020TrainingGA}, which we found beneficial for image quality and diversity. 
Considering the provided segmentation mask, we modify the discriminator feature augmentation (FA), ensuring that it does not interfere with the learning of the appearance of foreground objects. For this, the content FA is applied only to the representation of the background, while for the layout FA, the mixed spatial areas are sampled respecting the object boundaries in the segmentation mask. In our experiments, we observed this to be beneficial for the visual quality of images, as the model learnt to preserve the objects' appearance better.

\begin{table*}[b]
	
	\setlength{\tabcolsep}{0.25em}
	\renewcommand{\arraystretch}{1.12}
	\centering
	\centering
	\begin{tabular}{l|l l|l l}
		 Operation  & Input  & Size &  Output  & Size  \tabularnewline
		\hline
		\hline
{ConvTransp2D} & \texttt{z} & (64,1,1)          & {\texttt{up\_0}}  & {(256,3,5)}\\ \hline
		{ResBlock-Up} & \texttt{up\_0}  & (256,3,5)   & {\texttt{up\_1}} & {(256,3,5)} \\ 
		\hline
		{ResBlock-Up} & \texttt{up\_1}  & (256,3,5)   & {\texttt{up\_2}} & {(256,6,10)} \\ 
		\hline
		{ResBlock-Up} & \texttt{up\_2}  & (256,6,10)   & {\texttt{up\_3}} & {(256,12,20)} \\ 
		\hline
		{ResBlock-Up} & \texttt{up\_3}  & (256,12,20)   & {\texttt{up\_4}} & {(256,24,40)} \\ \hline
		{ResBlock-Up} & \texttt{up\_4}  & (256,24,40)   & {\texttt{up\_5}} & {(256,48,80)} \\ \hline
        {ResBlock-Up} & \texttt{up\_5}  & (256,48,80)   & {\texttt{up\_6}} & {(256,96,160)} \\ \hline
		{ResBlock-Up} & \texttt{up\_6}  & (256,96,160)   & {\texttt{up\_7}} & {(128,192,320)} \\ \hline
		{ResBlock-Up} & \texttt{up\_7}  & (128,192,320)   & {\texttt{up\_8}} & {(64,384,640)} \\ \hline \hline
		Conv2D, TanH           & \texttt{up\_5} & (256,48,80)     & \texttt{image\_3}  & (3,48,80)\\\hline
		Conv2D, TanH           & \texttt{up\_6} & (256,96,160)     & \texttt{image\_2}  & (3,96,160)\\\hline
		Conv2D, TanH           & \texttt{up\_7} & (128,192,320)     & \texttt{image\_1}  & (3,192,320)\\\hline
		Conv2D, TanH           & \texttt{up\_8} & (64,192,320)     & \texttt{image\_0}  & (3,384,640)\\\hline

	\end{tabular}
	
	\vspace{2ex}

	\caption{The OSMIS generator. The configuration is presented for the input noise of size $(3\times 5)$ and the final resolution of $(640\times384)$.}

	\label{supp_table:g_arch} %

\end{table*}

\begin{table*}[t]
	\setlength{\tabcolsep}{0.25em}
\renewcommand{\arraystretch}{1.12}
\centering
\centering
	\begin{tabular}{l|l l|l l}
		 Operation  & Input  & Size &  Output  & Size  \tabularnewline
		\hline
		\hline
		\multicolumn{5}{c}{Low-level discriminator $D_{low-level}$} \\ \hline
		Conv2D & \texttt{image\_0} & (3,384,640)              & \texttt{feat\_0}  & (32,384,640)\\\hline 
		Conv2D & \texttt{image\_1} & (3,192,320)              & \texttt{feat\_1}  & (8,192,320)\\\hline 
		Conv2D & \texttt{image\_2} & (3,96,160)              & \texttt{feat\_2}  & (16,96,160)\\\hline 
		Conv2D & \texttt{image\_3} & (3,48,80)              & \texttt{feat\_3}  & (32,48,80)\\\hline 	
			
		{ResBlock-Down} & 
		\texttt{feat\_0} & (32,384,640) & {\texttt{down\_0}} & {(64,192,320)} \\ \hline
		
		\multirow{2}{*}{ResBlock-Down} & \texttt{down\_0} & (64,192,320) & \multirow{2}{*}{\texttt{down\_1}} & \multirow{2}{*}{(128,96,160)} \\
& \texttt{feat\_1} & (8,192,320) & & \\ \hline

		\multirow{2}{*}{ResBlock-Down} & \texttt{down\_1} & (128,96,160) & \multirow{2}{*}{\texttt{down\_2}} & \multirow{2}{*}{(256,48,80)} \\
& \texttt{feat\_2} & (16,96,160) & & \\ \hline

		\multirow{2}{*}{ResBlock-Down} & \texttt{down\_2} & (256,48,80) & \multirow{2}{*}{\texttt{F}} & \multirow{2}{*}{(256,24,40)} \\
	& \texttt{feat\_3} & (32,48,80) & & \\ \hline \hline

		\multicolumn{5}{c}{Object discriminator $D_{object}$} \\ \hline
		MCA & \texttt{F} & (256,24,40)   & \texttt{F\_con} & N$\times$(256,1,1) \\\hline
		ResBlock-Down & \texttt{F\_con} & N$\times$(256,1,1)   & \texttt{cont\_0} & N$\times$(256,1,1) \\\hline
		ResBlock-Down & \texttt{cont\_0} & N$\times$(256,1,1)   & \texttt{cont\_1} & N$\times$(256,1,1) \\\hline
		ResBlock-Down & \texttt{cont\_1} & N$\times$(256,1,1)   & \texttt{cont\_2} & N$\times$(256,1,1) \\\hline
		ResBlock-Down & \texttt{cont\_2} & N$\times$(256,1,1)     & \texttt{cont\_3} & N$\times$(256,1,1) \\ \hline \hline
		\multicolumn{5}{c}{Layout discriminator $D_{layout}$} \\ \hline
		Conv2D & \texttt{F} & (256,24,40)   & \texttt{F\_lay} & (1,24,40) \\\hline
	ResBlock-Down & \texttt{F\_lay} & (1,24,40)   & \texttt{lay\_0} & (1,12,20) \\\hline
	ResBlock-Down & \texttt{lay\_0} & (1,12,20)   & \texttt{lay\_1} & (1,6,10) \\\hline
	ResBlock-Down & \texttt{lay\_1} & (1,6,10)   & \texttt{lay\_2} & (1,3,5) \\\hline
	ResBlock-Down & \texttt{lay\_2} & (1,3,5)     & \texttt{lay\_3} & (1,3,5) \\ \hline \hline

	\end{tabular}

	\vspace{2ex}

	\centering
	\caption{The OSMIS discriminator. The configuration is presented for the input noise of size $(3\times 5)$ and the final resolution of $(640\times384)$.}

\label{supp_table:d_arch} %
\end{table*}

\end{document}